\documentclass[runningheads]{llncs}

\usepackage{eccv}

\usepackage{eccvabbrv}

\usepackage{graphicx}
\usepackage{booktabs}

\usepackage[accsupp]{axessibility}  

\usepackage{hyperref}

\usepackage{orcidlink}

\begin{document}

\title{Geospecific View Generation - Geometry-Context Aware High-resolution Ground View Inference from Satellite Views} 

\titlerunning{Geospecific View Generation}

\author{Ningli Xu\orcidlink{0000-0003-0421-7349} \and
Rongjun Qin*\orcidlink{0000-0002-5896-1379}}

\authorrunning{Xu and Qin}

\institute{The Ohio State University, Columbus, OH 43210, USA \\
\email{\{xu.3961,qin.324\}@osu.edu}}

\maketitle

\begin{abstract}
Predicting realistic ground views from satellite imagery in urban scenes is a challenging task due to the significant view gaps between satellite and ground-view images. We propose a novel pipeline to tackle this challenge, by generating geospecifc views that maximally respect the weak geometry and texture from multi-view satellite images. Different from existing approaches that hallucinate images from cues such as partial semantics or geometry from overhead satellite images, our method directly predicts ground-view images at geolocation by using a comprehensive set of information from the satellite image, resulting in ground-level images with a resolution boost at a factor of ten or more. We leverage a novel building refinement method to reduce geometric distortions in satellite data at ground level, which ensures the creation of accurate conditions for view synthesis using diffusion networks. Moreover, we proposed a novel geospecific prior, which prompts distribution learning of diffusion models to respect image samples that are closer to the geolocation of the predicted images. We demonstrate our pipeline is the first to generate close-to-real and geospecific ground views merely based on satellite images. Code and dataset are available at \url{https://gdaosu.github.io/geocontext/}.
\end{abstract}
\let\thefootnote\relax\footnotetext{* corresponding author}
\let\thefootnote\relax\footnotetext{Pre-print version: to be published in ECCV 2024}

\section{Introduction}
\label{sec:intro}
The growing availability of satellites offers the opportunity to capture images in every corner of the world. Directly predicting ground-view images from these images, referred to as the cross-view synthesis problem, can benefit numerous applications, such as 3D realistic gaming \cite{li2024sat2scene}, and city-scale scene synthesis \cite{lu2020geometry,xu2024multi}. 

The primary challenges lie in significant disparities in viewing directions and resolutions across satellite and ground-level domains. Firstly, the difference in viewing angles makes the transformation from one view to the other very difficult and sensitive to noises. In urban areas, satellite images may capture subtle details of building facades. Transforming the visible facades to ground-view domains becomes highly sensitive to localization errors of building corners. Secondly, the low resolution of satellite images makes the extraction of useful information for ground-view synthesis difficult. The resolution of commercial satellite imagery is usually 0.3m/pixel, whereas the resolution of Google street-view imagery is much higher than satellite imagery, with around 3cm/pixel \cite{huang2022evaluation, xu2023point}. Bridging this nearly 10 times resolution difference remains a challenge, which cannot be simply addressed by super-resolution techniques. Finally, due to the diffuse reflections of clouds, the color distortions between satellite imagery and ground-view images are significant. 

\begin{figure}[tb]
    \centering
    \includegraphics[width=0.85\linewidth]{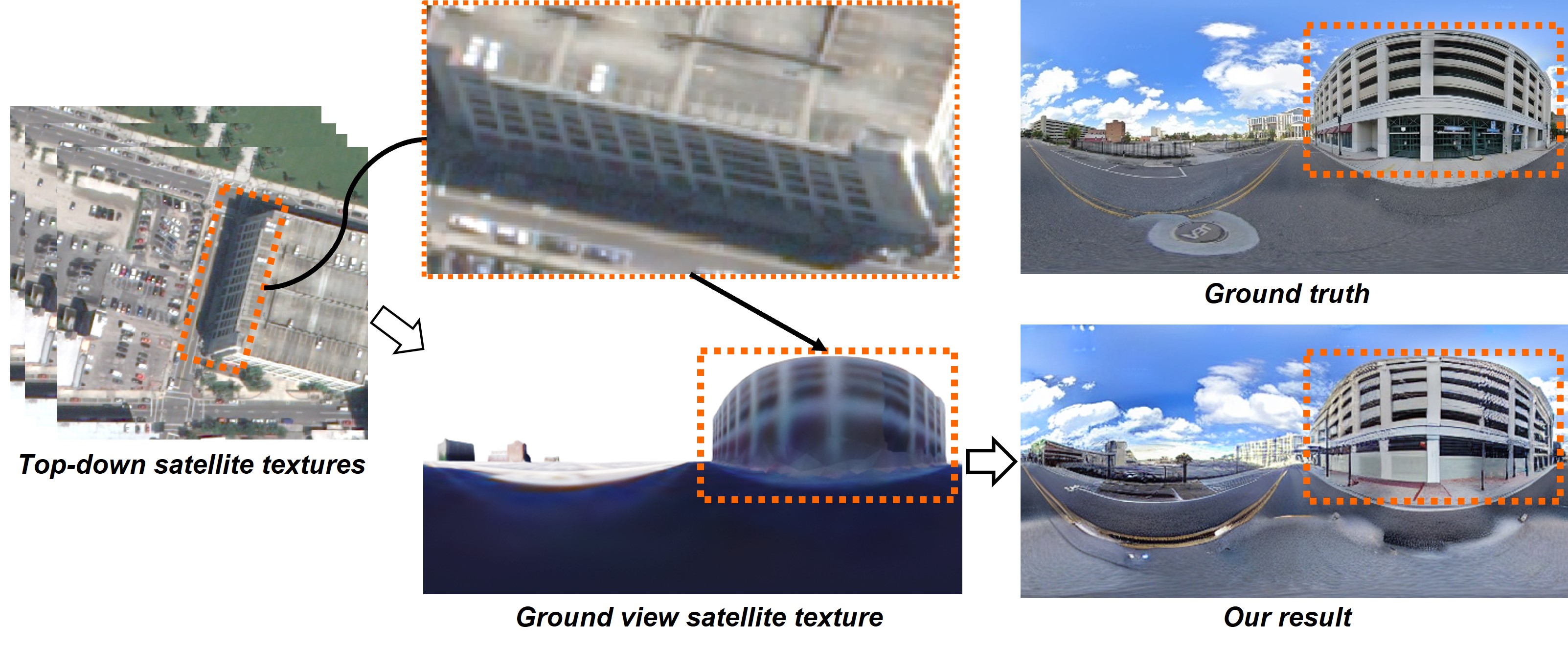}
    \caption{Example of our synthesized geospecific views. Instead of conditioning on semantics \cite{regmi2018cross,ren2021cascaded,lu2020geometry}, ours utilizes ground-view satellite texture which provides high-frequency structural and color information. The predicted result not only shows photorealistic quality but also accurately reflects the number of stories of the garage (marked as orange rectangles).}
    \label{fig:intro}
\end{figure}

Existing approaches \cite{ren2021cascaded,wu2022cross,lu2020geometry,li2021sat2vid,regmi2018cross} to address these challenges mainly seek for solutions that hallucinate views that reflect possible looks on the ground, which lack ground fidelity. They often adopt a black-box methodology or rely on auxiliary information. In tackling the disparity in viewing directions, \cite{wu2022cross,regmi2018cross} proposed end-to-end networks that directly learned the mapping relation between top-down satellite and ground-view images. As a result, the synthesized results lack photorealism and consistency in building facade regions due to substantial domain differences. \cite{lu2020geometry, li2021sat2vid} bridges the viewing direction difference by leveraging known accurate geometry. They proposed a 2D-3D-2D projection method that first projects the top-down satellite texture into 3D space via orthographic projection and then projects the 3D satellite texture into ground-view 2D space by panoramic projection. Nonetheless, orthographic projection is a compromised solution that sacrifices facade information. To address resolution differences, the majority of existing works \cite{ren2021cascaded,lu2020geometry,regmi2018cross,li2021sat2vid} employ cGAN-based methods \cite{choi2018stargan,isola2017image}, conditioning on ground-view semantics. However, this conditioning sacrifices texture information, making the synthesized results often deviate from the ground truth images.

Instead, our goal is to achieve ground view synthesis with not only photorealism but also maximal ground fidelity, meaning that the generated view will be geospecific, reflecting the actual looks at its geolocation. Our approach addresses the existing challenges for ground-view generation with a mathematically more accurate approach, avoiding any compromise of satellite texture and geometry information. Specifically, we proposed a novel cross-view synthesis approach with full utilization of satellite texture information including the visible building facades, as shown in \cref{fig:intro}. The projection from the top-down satellite to ground level is performed in the 2D-3D-2D way similar to \cite{lu2020geometry,li2021sat2vid} while the difference is the utilization of an accurate satellite camera projection model (rational polynomial camera, RPC \cite{singh2008rational}). The noises of the satellite geometry usually cause the distortions of projected texture. We developed a texture-friendly geometry refinement method to minimize distortions of the projected satellite texture. Additionally, we present a geospecific prior approach to improve the training efficiency and synthesis quality. Experimental results demonstrate that all baseline methods utilizing such textures exhibit superior synthesis quality compared to those relying on semantics. Our synthesized results not only excel in all perceptual metrics but also accurately capture building facade layouts. We summarize the main contributions of this work as:

\begin{itemize}
    \item The introduction of a texture-guided cross-view synthesis approach, which generates layout-preserving ground-view images conditioning building facade information.
    \item The development of a texture-friendly geometry refinement method allowing the utilization of subtle building facade details as the condition for cross-view synthesis. 
    \item Through rigorous experiments, we demonstrate our method outperforms SOTA methods at various metrics including semantic resemblance, edge, and perception similarity. 
\end{itemize}

\section{Related Work}
\label{sec:related}
\textbf{Cross-view synthesis} focuses on the novel view synthesis of objects or scenes from a completely different view. A typical task is to synthesize ground views given top-down view satellite images. Its main challenges are the huge view-port and domain difference. Existing works bridge such differences by using the top-down view or extracting high-level features from the top-down view as the condition for the ground-view synthesis. \cite{regmi2018cross,wu2022cross} used conditional GANs \cite{isola2017image} predicting both the ground-view and corresponding semantics conditioning on top-down view satellite image. The large view-port difference makes such methods difficult to converge. Instead, \cite{lu2020geometry, qian2023sat2density, li2021sat2vid, li2024sat2scene} perform viewport transformation based on predicted geometry, where they estimated the height maps from top-down views assuming the orthographic projection. As an approximation projection, it can preserve the roof and ground information while ignoring the building facade information. Our method developed a robust way of transforming such information to ground views while preserving the geometric consistency, which further served as the synthesis condition.

\textbf{Conditional image generation} focuses on learning a parametric mapping between source condition domains and target image domains. Example conditions include text descriptions to generate corresponding images \cite{rombach2022high, reed2016learning}, broken images to fill the missing parts \cite{pathak2016context,lugmayr2022repaint, rombach2022high}, street-view semantics to generate the corresponding images \cite{isola2017image, zhu2017unpaired, zhang2023adding}. Among these works, conditional GANs \cite{choi2018stargan,isola2017image} are widely used as the backbone for conditional image generation while they suffer from slow and unstable convergence during the training process and require a large volume of paired data. Recently, diffusion models \cite{ho2020denoising, song2020score, rombach2022high} have proven exceptional in image generation tasks, which iteratively denoises the Gaussian noise distribution to the target image distribution. Compared to GANs, they are more adapted to various image domains \cite{zhang2023adding,ruiz2022dreambooth} with the limited amount of training data \cite{roich2022pivotal,ruiz2022dreambooth,hu2021lora}. Our method explores a way of ground-view synthesis conditioning on a combination of geo-location and context information using ControlNet \cite{zhang2023adding} and LORA \cite{hu2021lora}.

\textbf{Cross-view geo-localization} It focuses on estimating the location and orientation of ground-view images based on given satellite images. Early works regard it as the image retrieval problem that finds the most similar satellite image from a database to determine the rough location of query images. These works focus on designing powerful handcrafted features \cite{lin2013cross,castaldo2015semantic} or learning-based features \cite{cai2019ground,vo2016localizing,workman2015wide,regmi2019bridging} to bridge the cross-view domain gap. \cite{zhu2021vigor,Hu_2018_CVPR} further identified the accurate pixel location on satellite images corresponding to the query images by employing the Siamese network to regress the location coordinates. Recent works \cite{shi2023boosting, lentsch2023slicematch} utilized the geometry guidance to project the ground-view images to the top-down view domain, where the location and orientation of query images can be robustly regressed. 

\begin{figure}[tb]
    \centering
    \includegraphics[width=\textwidth]{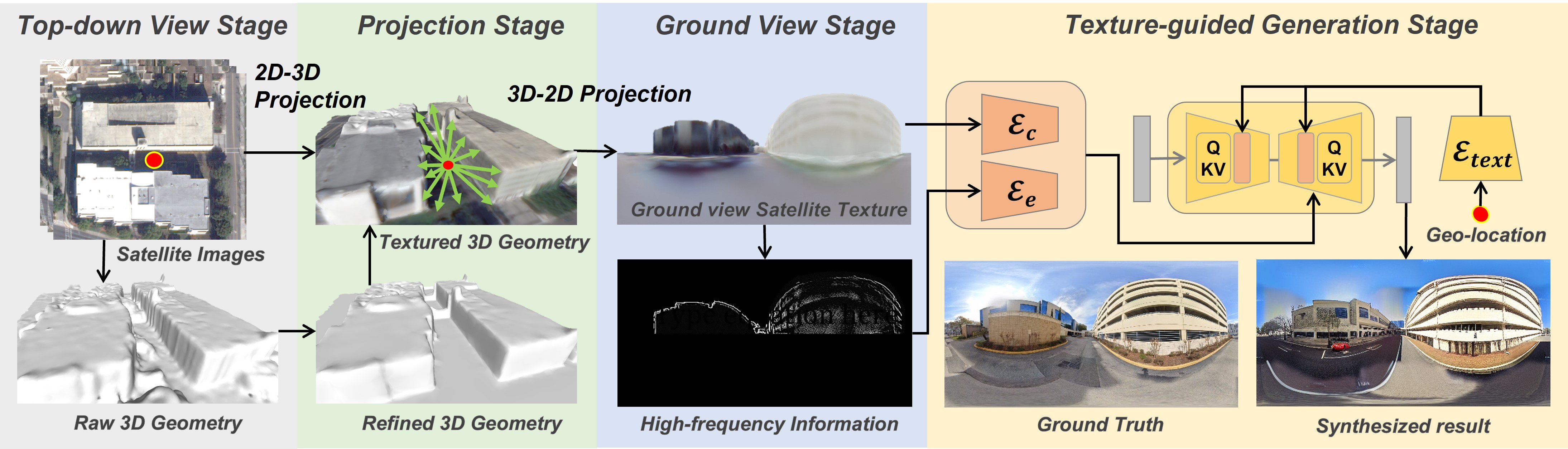}
    \caption{Overview of our pipeline. \textbf{Top-down View Stage} and \textbf{Projection Stage}: the satellite textures are projected to the refined 3D geometry and then projected back to ground-view 2D space (\cref{sec:sec31}). \textbf{Ground-view Stage:} The ground view satellite texture and corresponding high-frequency layout information serve as the conditions (\cref{sec:sec32}). \textbf{Texture-guided Generation Stage:} We use the recent successful diffusion model \cite{rombach2022high} conditioning on ground-view satellite textures, high-frequency information with the geospecific prior. (\cref{sec:sec33})}.
    \label{fig:pipeline}
\end{figure}

\section{Method}
\label{sec:sec3}
We present a novel pipeline designed to predict ground-view panorama images using a set of satellite images, as depicted in \cref{fig:pipeline}. Our main goal is to perform geometrically accurate projection of satellite textures to the ground view, encompassing subtle details of building facades to enhance ground-view synthesis. The proposed pipeline consists of four stages: the top-down view stage, projection stage, ground-view stage, and texture-guided generation stage. The details are described below. 

\subsection{Top-down View and Projection Stage}
\label{sec:sec31}

We follow the 2D-3D-2D way to project the top-down satellite images to ground level. The satellite 3D geometry is first derived via well-established stereo matching methods \cite{de2014automatic,Leotta_2019_CVPR_Workshops,RPC, xu2024large}. We then perform a 2D-3D texture projection (known as RPC projection \cite{singh2008rational}) to transform the top-down satellite textures to 3D space and fuse the multiple overlap textures by optimizing the global illumination and color consistency. Then a panoramic projection is performed from the 3D texture information to the ground level. The crucial factor in achieving perfect 2D-3D projection lies in the precise and smooth geometry. Although the derived 3D geometry is mathematically computed based on multi-view constraints, the presence of satellite sensor noises can introduce distortions that will largely impact the quality of projected textures around building facades. Therefore, we propose an effective approach to refine the geometry of satellite buildings.

\begin{figure}[tb]
    \centering
    \includegraphics[width=0.85\linewidth]{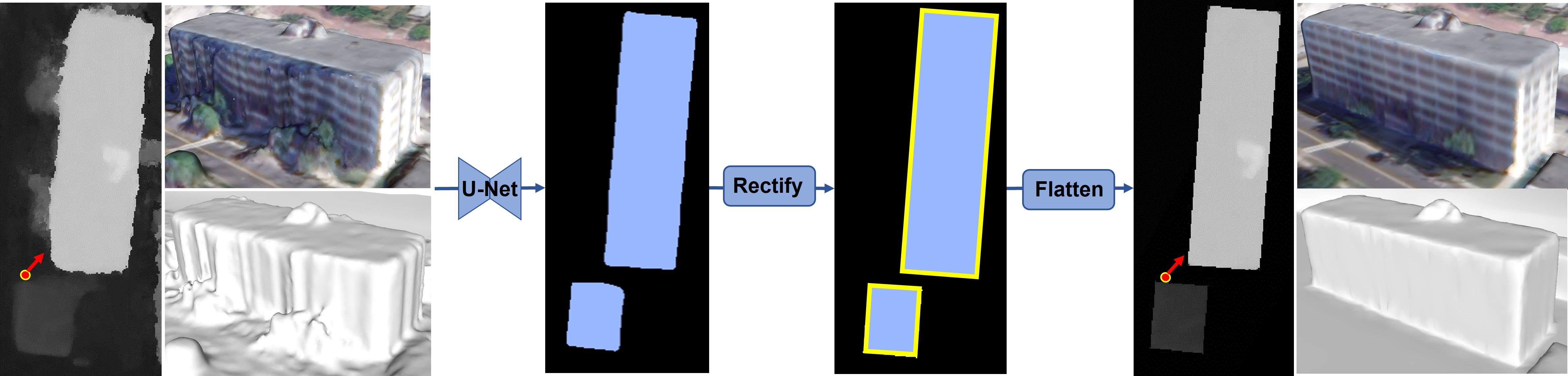}
    \caption{Texture-friendly geometry refinement process. The process takes the original height map as input and estimates the building footprint, followed by boundary regularization to produce the refined height map.}
    \label{fig:rectify_pipeline}
\end{figure}

\begin{figure}[tb]
    \centering
      \captionsetup[subfigure]{justification=centering}
      \begin{subfigure}[t]{0.85\linewidth}
      \includegraphics[width=\linewidth]{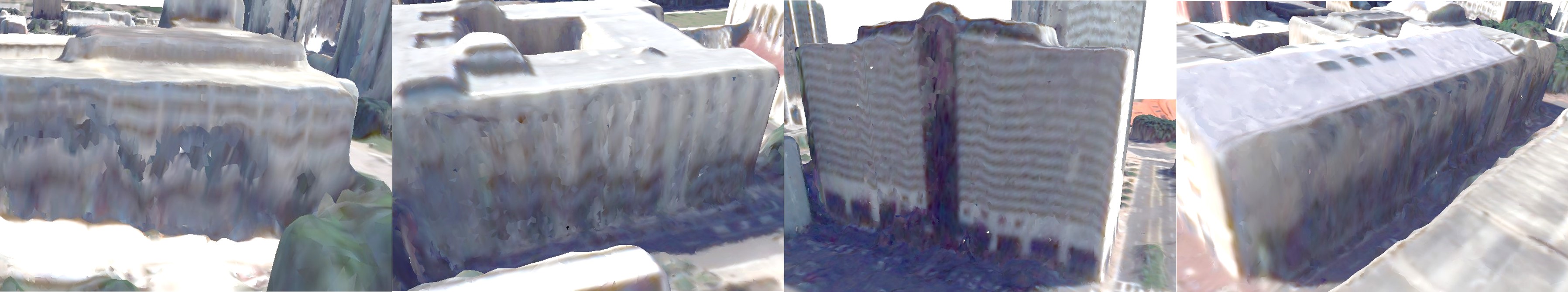}
      \caption{Before refinement}
      \end{subfigure}
            \captionsetup[subfigure]{justification=centering}
      \begin{subfigure}[t]{0.85\linewidth}
      \includegraphics[width=\linewidth]{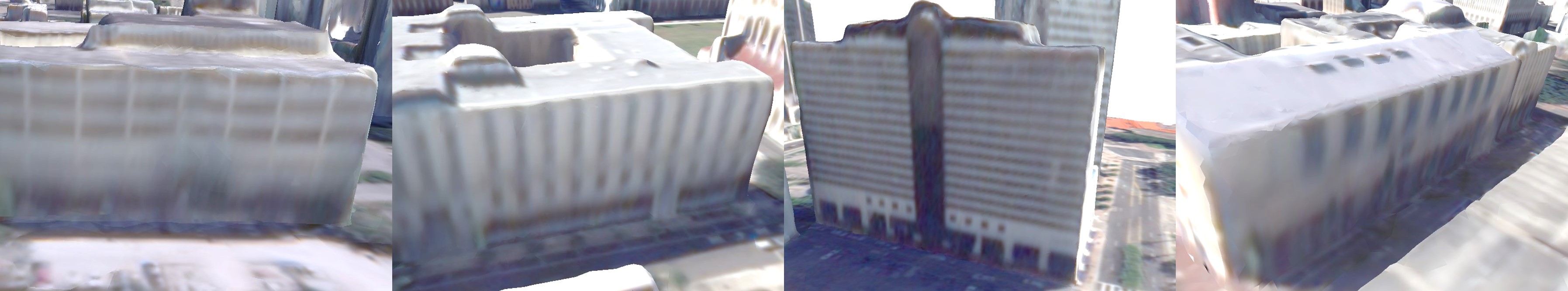}
      \caption{After refinement}
      \end{subfigure}
    \caption{Examples of the satellite textures before and after our transformation-friendly geometry refinement.}
    \label{fig:refine_result}
\end{figure}

\textbf{Texture-friendly geometry refinement}. As the building facades are nearly perpendicular to the satellite viewing direction, even minor disturbances around the facade surface can lead to inaccuracies in mapping satellite textures to the correct facade location. To ensure the smoothness and precision of the satellite building geometry, we employ a 2D U-Net \cite{xie2021segformer} to ascertain the high-level building footprint from satellite images, classifying each pixel as either part of a building or non-building region, as shown in \cref{fig:rectify_pipeline}. Subsequently, we detect and rectify building boundaries into a series of polygons, which will provide smooth building boundaries. For non-building regions, we conduct plane fitting and flatten the non-building pixels onto the fitted plane, while retaining the original configuration of building pixels. Once the satellite geometry is refined, we perform 2D-3D texture projection, and then panoramic projection to derive the ground-view satellite textures. Our refinement method shows excellent results on various buildings, examples as shown in \cref{fig:refine_result}.

\subsection{Ground-view Stage}
\label{sec:sec32}
After minimizing the cross-viewport difference between satellite and ground-view images, the remaining challenges primarily involve resolution and color, with the resolution exhibiting a difference of over 10 times. To address this resolution disparity, we introduce a novel texture-guided condition to enhance the informativeness of ground-view generation.

\begin{figure}[tb]
    \centering
      \begin{subfigure}[t]{0.3\linewidth}
    \includegraphics[width=\linewidth]{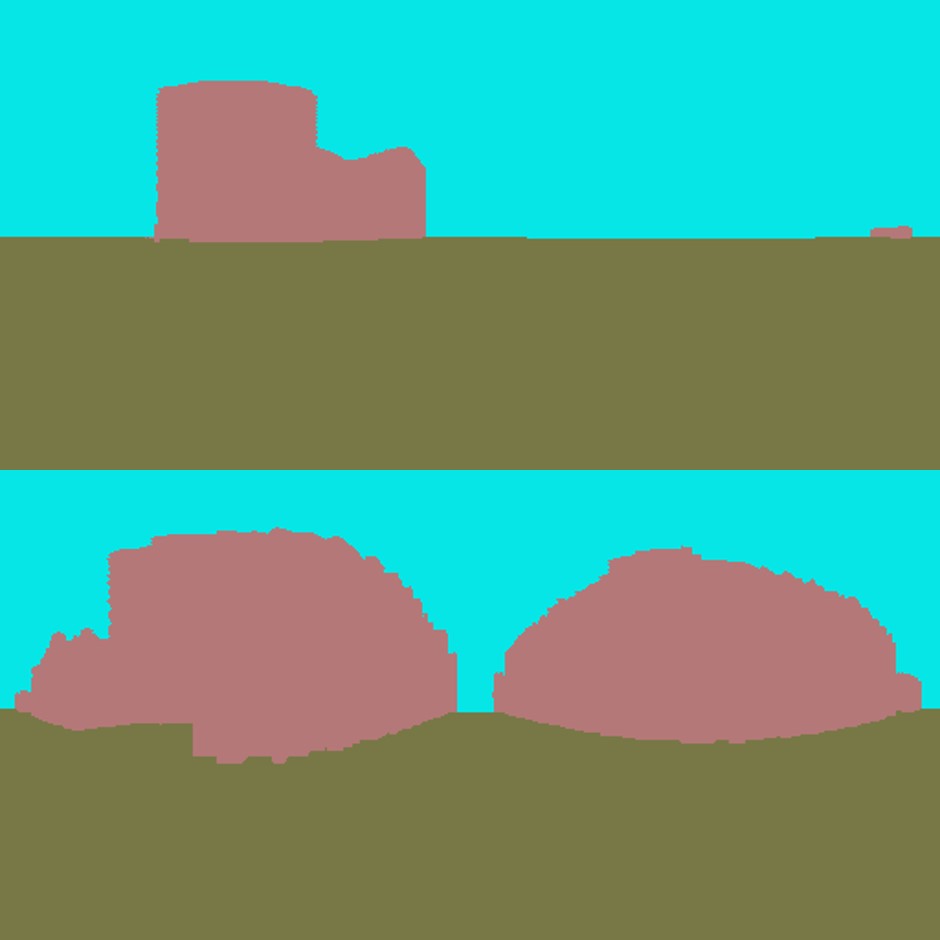}
    \caption{Semantics}
  \end{subfigure}
        \begin{subfigure}[t]{0.3\linewidth}
    \includegraphics[width=\linewidth]{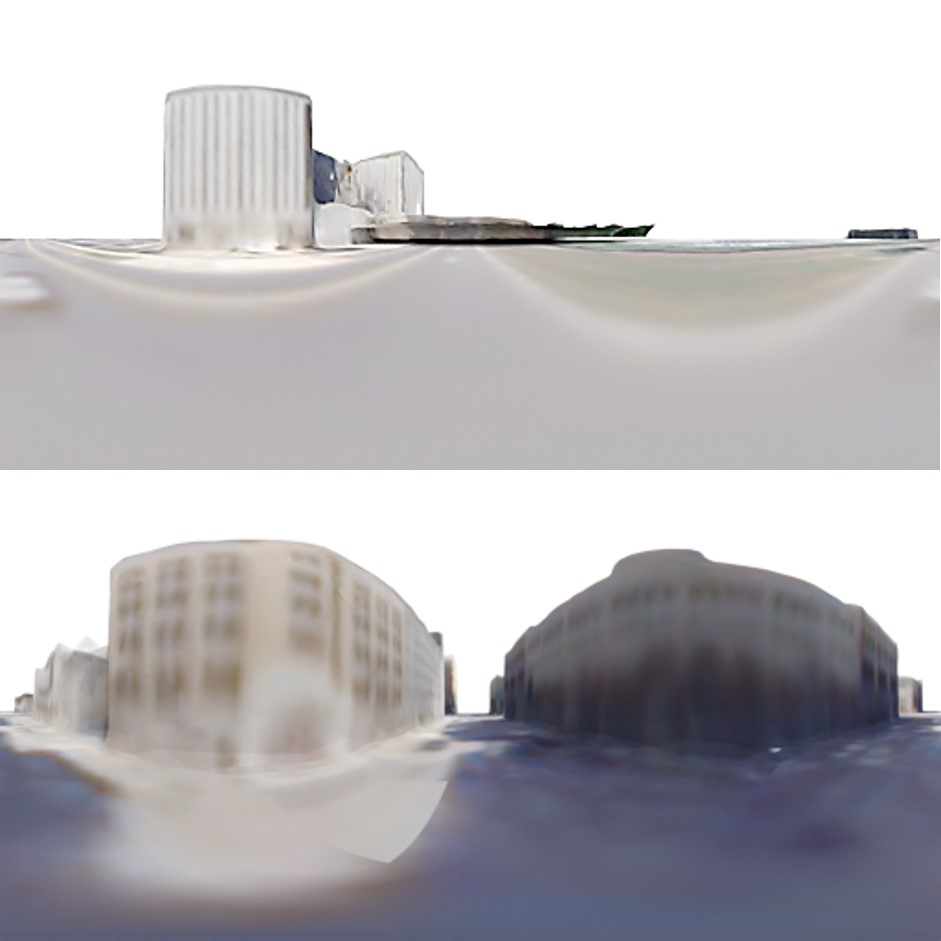}
    \caption{Our satellite texture}
  \end{subfigure}
        \begin{subfigure}[t]{0.3\linewidth}
    \includegraphics[width=\linewidth]{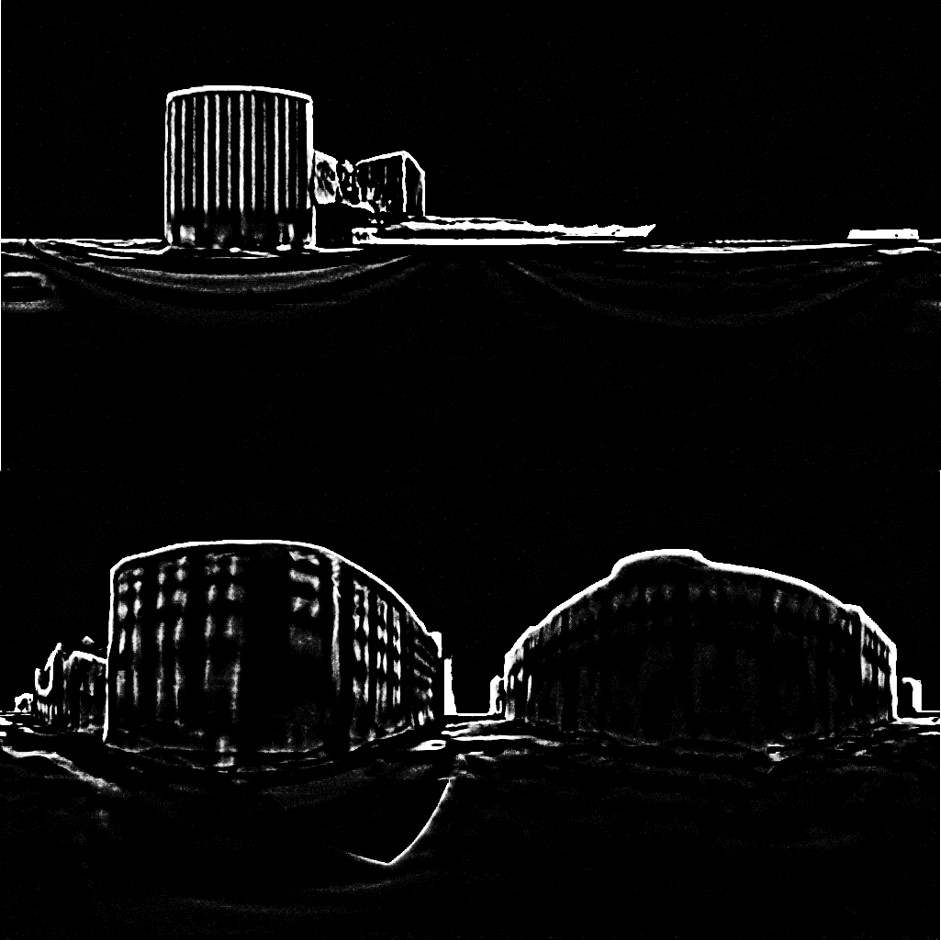}
    \caption{Our edge map}
  \end{subfigure}
    \caption{Illustration of three conditions for cross-view synthesis. Semantics are widely used by existing works \cite{wu2022cross,castaldo2015semantic,lu2020geometry}. Our satellite textures can provide additional high-frequency and color information that details the building facade layouts, such as the window/door shape and locations.}
    \label{fig:condition}
\end{figure}

\textbf{Texture-guided condition}. Most existing cross-view synthesis works \cite{wu2022cross,castaldo2015semantic,lu2020geometry} conditions on semantics, which are assumed as the known information \cite{castaldo2015semantic}, internally estimated \cite{wu2022cross} or by 2D-3D-2D projection \cite{lu2020geometry}. Deriving the semantics from satellite imagery is difficult and the semantic labels are limited to certain classes such as "Building", "Sky", "Road/Ground" and "Trees", limiting the diversity and richness of the details. Our semantics are generated based on top-down view building footprint segmentation results and perform 2D-3D-2D projection detailed in \cref{sec:sec31}. In addition, we extract non-categorical, high-frequency information to preserve small granular structural details. Utilizing a 2D U-Net \cite{chan2022learning}, we extract the building facade layout information, as shown in \cref{fig:condition}.

\subsection{Texture-guided Generation Stage}
\label{sec:sec33}

We apply the latent diffusion model \cite{rombach2022high} as the base generator, which iteratively performs the denoising process from 2D random noise maps to synthesize the ground-view images. 

\begin{figure}[tb]
    \centering
    \includegraphics[width=0.85\linewidth]{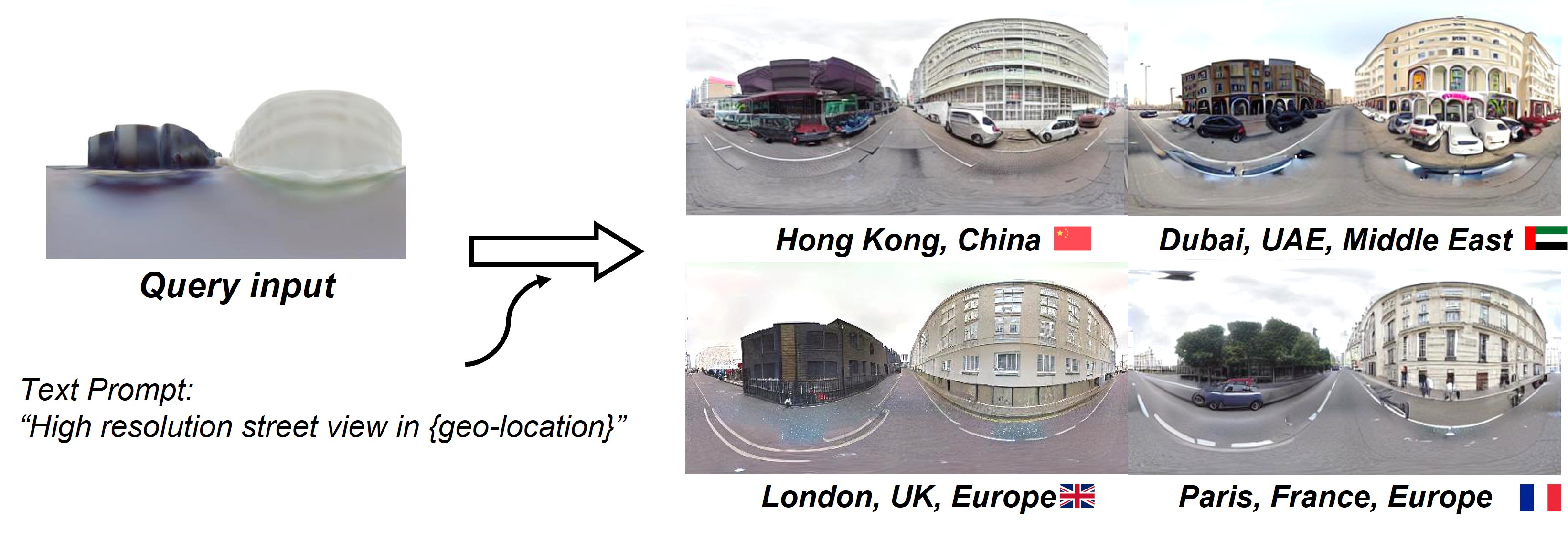}
    \caption{Examples of the synthesized results of different geospecific priors conditioning on the same query condition.}
    \label{fig:fig1}
\end{figure}

\textbf{Geospecific information prior}. Many small-scale geographical nuances, such as specific vegetation types (e.g., palm trees in tropical regions), and diverse building facade features (including billboards and neon lights in Hong Kong), cannot be adequately captured by pure satellite texture. To address this, we embed the geospecific prior as the additional learnable parameters \cite{hu2021lora, ruiz2023dreambooth} to the cross-attention module of our diffusion model. The geospecific information is represented as text descriptions corresponding to specific countries or regions, aligned with a set of street-view images from those areas to present the typical landscape of such regions. Specifically, given a geo-specific text prompt "High resolution street view in \{geospecific\}" \(\mathbf{P}\), we obtain a conditioning vector \(\mathbf{c}_p = \mathcal{E}_{text}(\mathbf{P})\). Subsequently, we incorporate geo-location tokens \(\mathbf{P}_t\)  (e.g. HongKong, Dubai, Paris) into the prompt \(\mathbf{P}\) to  produce geospecific conditioning vector \(\mathbf{c}_t\), formalized as:
\begin{equation}\label{eq:prior_feature}
    \mathbf{c}_t = \mathcal{E}_{text}(\mathbf{P} + \mathbf{P}_t)
\end{equation}
For the pre-trained weights of our diffusion model \(W_0 \in R^{d \times k}\), where \(k\) is the dimension for the input feature vectors and \(d\) is the dimension of the output feature vectors, we follow \cite{hu2021lora} to introduce two low-rank matrices \(A \in \mathbb{R}^{d \times r}, B \in \mathbb{R}^{r \times k}\) and rank \(r \ll min(d,k)\).
During the training process, using random Gaussian initialization for \(A\) and zero initialization for \(B\), and the optimizer only optimizes these two matrices, the output \(h\) pass pre-trained weight \(W_0\) will be modified as:
\begin{equation}
    h = W_0x + ABx
\end{equation}
where \(x\) is the feature vector in the diffusion model. \cref{fig:fig1} illustrates that our method with geo-specific priors can generate images with geospecific attributes.

\textbf{Texture Encoding}. The ground-view satellite texture and edge map are encoded by another two networks \cite{zhang2023adding} sharing the same architecture as our diffusion model, which conditions on such features by zero-convoluting them with each layer of the base model decoder. Given an input pair of images \(\{\mathbf{z}_0, \mathbf{c}\}\), where \(\mathbf{z}_0\) is the real street-view image, \(\mathbf{c}\) are ground-view satellite textures. We first convert the condition images into feature space following the VQ-GAN \cite{Esser_2021_CVPR} pre-processing pipeline.
\begin{equation}
    \mathbf{c}_s = \mathcal{E}(\mathbf{c}).
\end{equation}
where \(\mathbf{c}_s\) is the texture condition vector that represents the color information of the satellite image. To incorporate the building structure information, we extract the edge condition vector \(\mathbf{c}_e\) from \(\mathbf{c}\)
\begin{equation}
    \mathbf{c}_e = \mathcal{E}(\mathcal{H}(\mathbf{c})).
\end{equation}
where \(\mathcal{H}\) is the edge map extraction network described in \cref{sec:sec32}.

During the training process, given a time step \(\mathbf{t}\) and a geospecific prompt (encoded as feature vector \(\mathbf{c}_t\), see \cref{eq:prior_feature}), our diffusion-based network progressively adds Gaussian noise \(\epsilon \sim \mathcal{N}(0,1)\) to the previous image \(\mathbf{z}_{t-1}\) and produces a new noisy image \(\mathbf{z}_t\) and it learns to predict the noise by minimizing the mean-square error:
\begin{equation}
    \mathcal{L} = \mathbb{E}_{z_0,t,\mathbf{c}_s,\mathbf{c}_e,\mathbf{c}_t \epsilon \sim \mathcal{N}(0,1)} \left[ \left\| \epsilon - \epsilon_{\theta}(z_t, t, \mathbf{c}_s, \mathbf{c}_e, \mathbf{c}_t) \right\|_2^2 \right]
\end{equation}
Where \(\mathcal{L}\) is the learning objective applied in our proposed approach. We aim to finetune the two texture encoding networks, where the first is conditioned by ground-view satellite RGB images, and the second is conditioned by the edge map.
\begin{figure}[tb]
    \centering
    \includegraphics[width=0.85\linewidth]{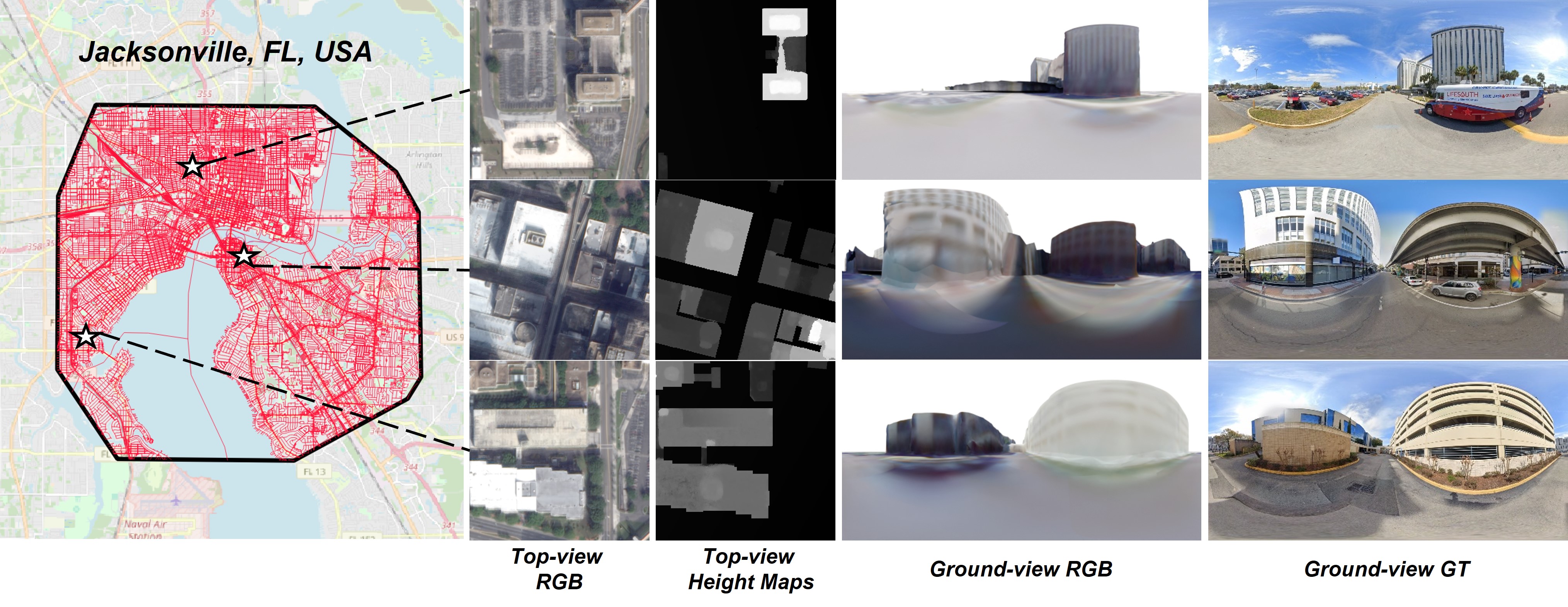}
    \caption{Examples of our cross-view dataset based on Jacksonville, Florida, USA. It provides well-aligned pairs of ground-view satellite RGB and ground truth images, along with the top-view RGB and height maps.(see details in \cref{sec:dataset})}
    \label{fig:dataset}
\end{figure}
\subsection{Implementation Details}
\textbf{Datasets}.\label{sec:dataset}
We perform our experiments on a large-scale dataset, DFC 2019 \cite{dfc2019}, consisting of multi-view satellite images covering a 177\(km^2\) area in Jacksonville, USA. Examples of satellite data and associated products are shown in \cref{fig:dataset}, based on which we process into the 3D models, and have collected corresponding ground-views from Google street-view. Specifically, we first conducted stereo matching \cite{hirschmuller2007stereo,RPC} and our texture-friendly geometry refinement as described in \cref{sec:sec31} to produce refined satellite geometry in the form of a height map. Ground-view depth maps were then generated through 3D-2D projection from the refined satellite geometry. Subsequently, we performed 2D-3D-2D projection to map the top-down satellite texture to ground level. Semantic information was derived from OpenStreetMap Building Footprint data\cite{OpenStreetMap}. To obtain the reference ground view data, we collected Google street-view images within the study area using Google StreetView 360, with a step distance of 30 meters. Each image included location information (longitude, latitude, orientation). To address positional errors, we adopted a preprocessing strategy similar to \cite{lu2020geometry}, calculating the overlap ratio between sky regions in Google street-view images and ground-view satellite images. We selected pairs with an overlap ratio exceeding 95\%, followed by post-processing. This process yielded over 7,000 pairs of cross-view data, each comprising top-down view RGB, height map, ground-view RGB, depth map, and ground truth images. For dataset split, we spatially tiled the datasets into a set of \(700m \times 700m\) sub-tiles, with the ratio of train/val/test tiles being \(8:1:1\). To prevent spatial correlation and overfitted prediction, we selected our test tile far apart from the training tiles.

\textbf{Training Scheme}. For the building footprint segmentation network, we randomly cropped out 512 x 512 patches from satellite images and assigned each pixel with 1 or 0 representing whether belongs to the building or not based on OpenStreetMap data with post-processing. This building footprint dataset is self-constructed based on the DFC-2019 multi-view images, which contain 11,784 pairs. Given its moderate size and simplicity of the task (predicting a binary task), We chose a moderately complex network, SegFormer \cite{xie2021segformer} as our segmentation network, which was trained for 40 epochs.

For the texture-guide generation stage, we used pre-trained weights from Stable Diffusion v1.5 \cite{rombach2022high} as the base diffusion model and only finetuned the newly added parameters for each geospecific text prompt (detailed in \cref{sec:sec33}). The paired data for geospecific prior training was a text prompt of "High-resolution street view in <City, Country, Continent>" and a set of Google street-view images broadly in the region to reflect the types of buildings and city landscape. In our experiment, we encoded five cities (but easily expandable) including London, Hong Kong, Jacksonville, Paris, and Dubai, where each city contains around 500 images. For ground-view texture encoders, we created the trainable copy of the UNet encoder according to ControlNet \cite{zhang2023adding} and trained the encoders with the train datasets and inference on test datasets.
\begin{table}[tb]
  \centering
    \caption{Quantitative evaluation of synthesized image quality. We compared our method with  Sat2Ground \cite{lu2020geometry}, Sat2Density \cite{qian2023sat2density}, CrossMLP \cite{ren2021cascaded} and PanoGAN \cite{wu2022cross}. The same metrics for \cref{tab:quan_ablation}.}
\begin{tabular}{c c c c c c c c c c c c}
    \toprule
    \textbf{Method} & \multicolumn{2}{c}{\textbf{Low level}} & \textbf{Edge level} &\multicolumn{3}{c}{\textbf{Semantic level}} & \multicolumn{3}{c}{\textbf{Perceptual Level}} \\
 & PSNR\(\uparrow\) & SSIM\(\uparrow\) & \(I_E\)\(\uparrow\) & \(I_B\)\(\uparrow\) & \(I_G\)\(\uparrow\) & \(I_S\)\(\uparrow\) & LPIPS\(\downarrow\) & FID\(\downarrow\) & DreamSIM\(\downarrow\)\\ 
    \midrule
      Sat2G & \textbf{21.02} & 0.388 & 0.072 & 0.345 & 0.324 & 0.851 & 0.527 & 160.6 & 0.420\\  
  Sat2D & 19.04 & 0.388 & 0.067 & 0.285 & 0.310 & 0.782 & 0.574 & 227.3 & 0.481\\  
CrossMLP & 18.66 & \textbf{0.407} & 0.069 & 0.448 & 0.214 & 0.861 & 0.509 & 170.8 & 0.434\\  
 PanoGAN & 20.51 & 0.373 & 0.078 & 0.376 & 0.457 & 0.801 & 0.488 & 98.81 & 0.348\\  
  Ours & 19.95 & 0.397 & \textbf{0.089} & \textbf{0.570} & \textbf{0.864} & \textbf{0.874} & \textbf{0.449} & \textbf{71.04} & \textbf{0.315}\\ 
    \bottomrule
  \end{tabular}
  \label{tab:sota_quant}
\end{table}

\section{Experiments}
\label{sec:experiments}
\subsection{Baselines and Metrics}
\label{sec:baseline}
\def\sota_width{0.16}
\begin{figure}[tb]
  \captionsetup[subfigure]{justification=centering}
  \begin{subfigure}[t]{\sota_width\linewidth}
  \includegraphics[width=\linewidth]{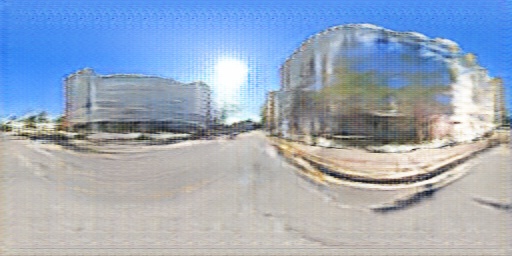}
  \end{subfigure}
    \begin{subfigure}[t]{\sota_width\linewidth}
  \includegraphics[width=\linewidth]{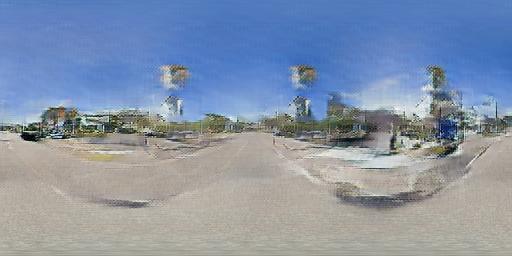}
  \end{subfigure}
  \begin{subfigure}[t]{\sota_width\linewidth}
  \includegraphics[width=\linewidth]{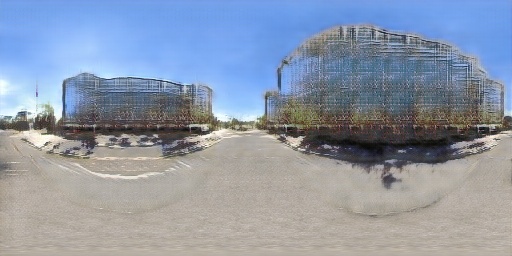}
  \end{subfigure}
  \begin{subfigure}[t]{\sota_width\linewidth}
  \includegraphics[width=\linewidth]{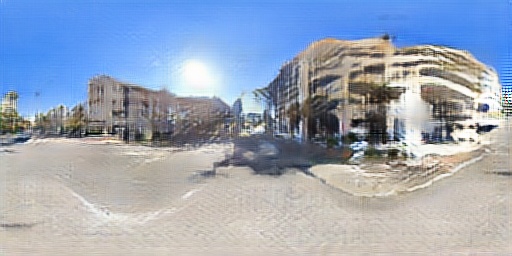}
  \end{subfigure}
  \begin{subfigure}[t]{\sota_width\linewidth}
  \includegraphics[width=\linewidth]{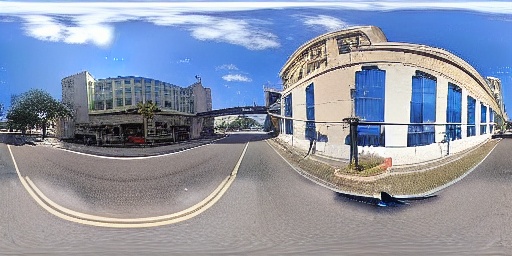}
  \end{subfigure}
  \begin{subfigure}[t]{\sota_width\linewidth}
    \includegraphics[width=\linewidth]{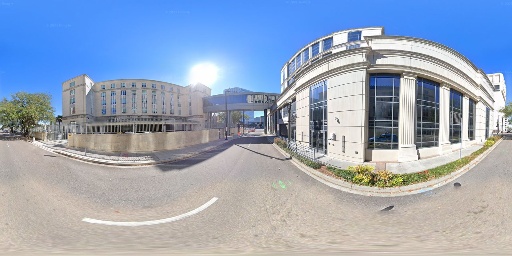}
  \end{subfigure}
  \begin{subfigure}[t]{\sota_width\linewidth}
  \includegraphics[width=\linewidth]{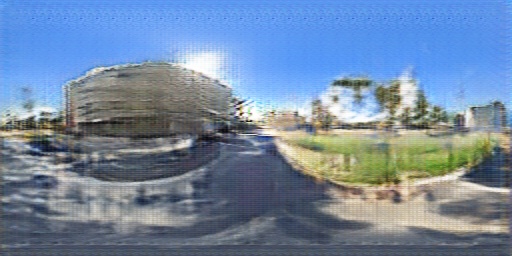}
  \end{subfigure}
  \begin{subfigure}[t]{\sota_width\linewidth}
  \includegraphics[width=\linewidth]{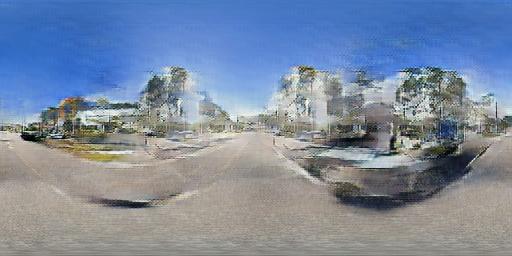}
  \end{subfigure}
  \begin{subfigure}[t]{\sota_width\linewidth}
    \includegraphics[width=\linewidth]{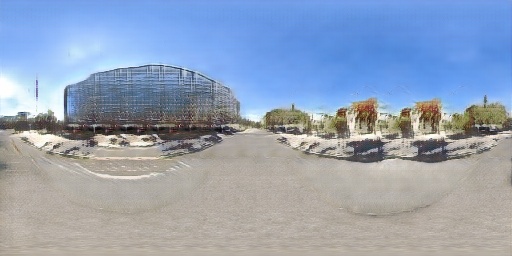}
  \end{subfigure}
  \begin{subfigure}[t]{\sota_width\linewidth}
    \includegraphics[width=\linewidth]{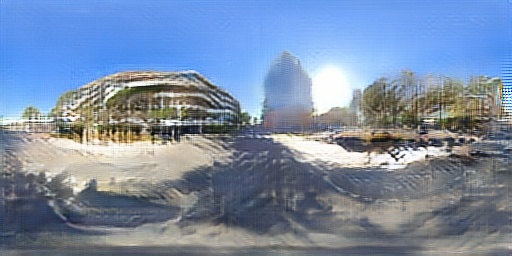}
  \end{subfigure}
  \begin{subfigure}[t]{\sota_width\linewidth}
    \includegraphics[width=\linewidth]{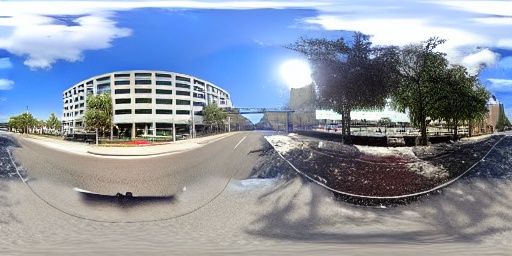}
  \end{subfigure}
  \begin{subfigure}[t]{\sota_width\linewidth}
  \includegraphics[width=\linewidth]{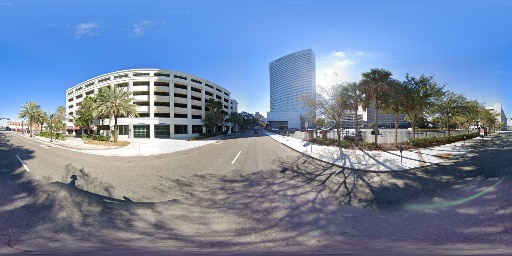}
  \end{subfigure}
    \begin{subfigure}[t]{\sota_width\linewidth}
  \includegraphics[width=\linewidth]{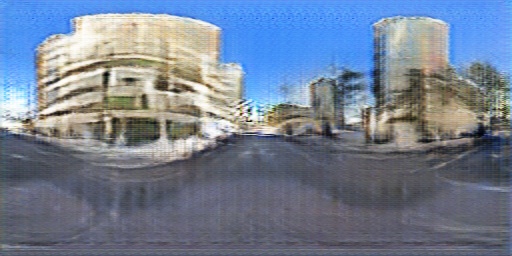}
  \end{subfigure}
  \begin{subfigure}[t]{\sota_width\linewidth}
  \includegraphics[width=\linewidth]{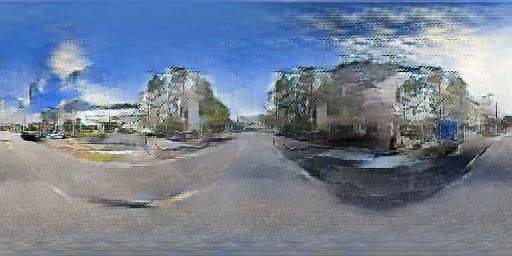}
  \end{subfigure}
  \begin{subfigure}[t]{\sota_width\linewidth}
    \includegraphics[width=\linewidth]{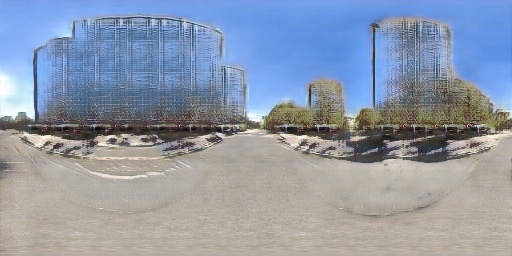}
  \end{subfigure}
  \begin{subfigure}[t]{\sota_width\linewidth}
    \includegraphics[width=\linewidth]{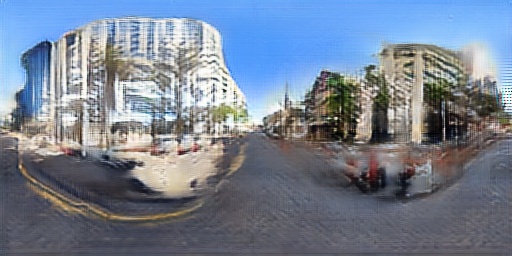}
  \end{subfigure}
  \begin{subfigure}[t]{\sota_width\linewidth}
    \includegraphics[width=\linewidth]{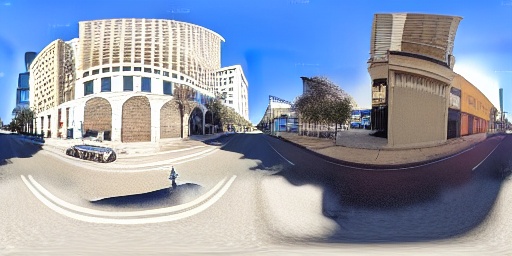}
  \end{subfigure}
  \begin{subfigure}[t]{\sota_width\linewidth}
  \includegraphics[width=\linewidth]{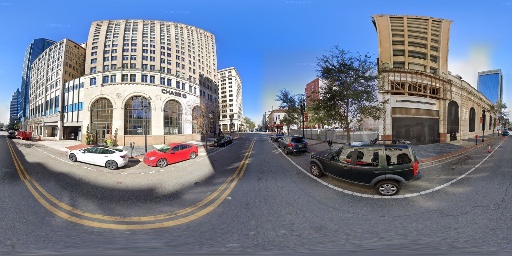}
  \end{subfigure}
    \begin{subfigure}[t]{\sota_width\linewidth}
  \includegraphics[width=\linewidth]{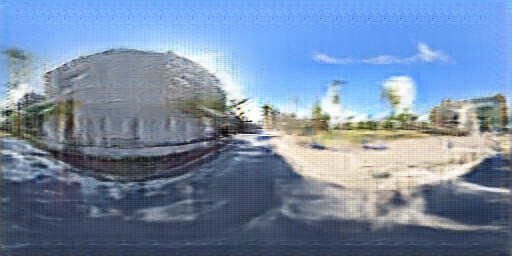}
  \end{subfigure}
  \begin{subfigure}[t]{\sota_width\linewidth}
  \includegraphics[width=\linewidth]{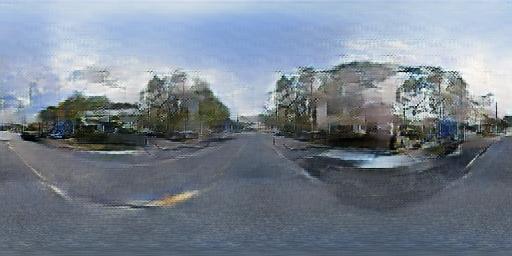}
  \end{subfigure}
  \begin{subfigure}[t]{\sota_width\linewidth}
    \includegraphics[width=\linewidth]{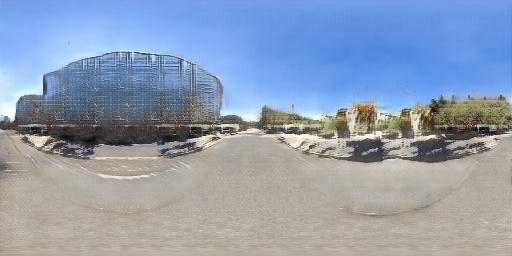}
  \end{subfigure}
  \begin{subfigure}[t]{\sota_width\linewidth}
    \includegraphics[width=\linewidth]{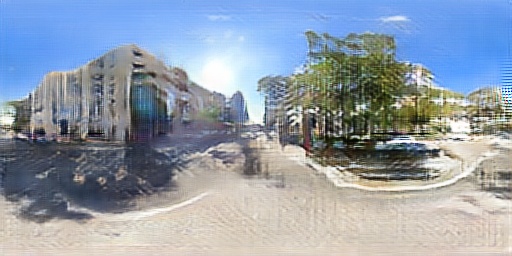}
  \end{subfigure}
  \begin{subfigure}[t]{\sota_width\linewidth}
    \includegraphics[width=\linewidth]{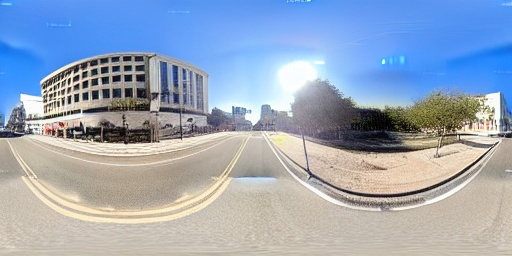}
  \end{subfigure}
  \begin{subfigure}[t]{\sota_width\linewidth}
  \includegraphics[width=\linewidth]{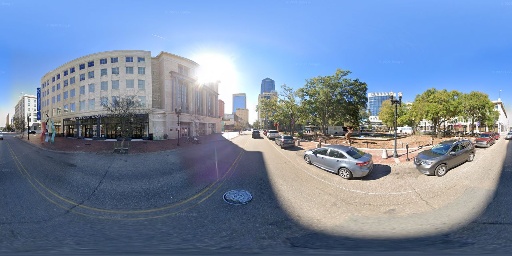}
  \end{subfigure}
    \begin{subfigure}[t]{\sota_width\linewidth}
  \includegraphics[width=\linewidth]{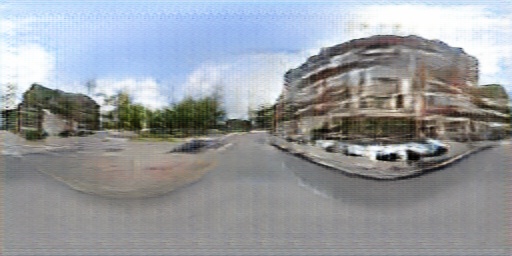}
  \end{subfigure}
  \begin{subfigure}[t]{\sota_width\linewidth}
  \includegraphics[width=\linewidth]{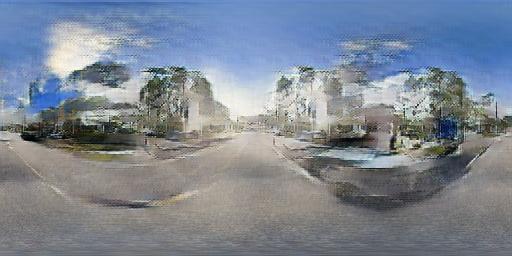}
  \end{subfigure}
  \begin{subfigure}[t]{\sota_width\linewidth}
    \includegraphics[width=\linewidth]{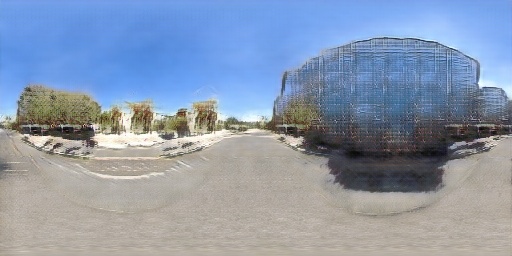}
  \end{subfigure}
  \begin{subfigure}[t]{\sota_width\linewidth}
    \includegraphics[width=\linewidth]{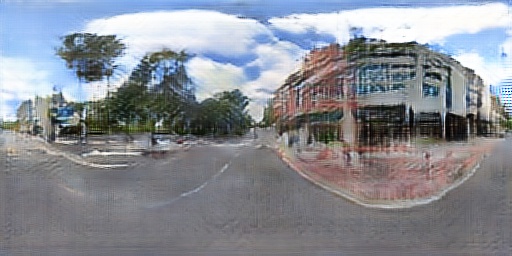}
  \end{subfigure}
  \begin{subfigure}[t]{\sota_width\linewidth}
    \includegraphics[width=\linewidth]{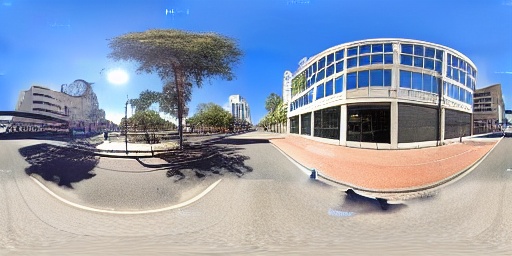}
  \end{subfigure}
  \begin{subfigure}[t]{\sota_width\linewidth}
  \includegraphics[width=\linewidth]{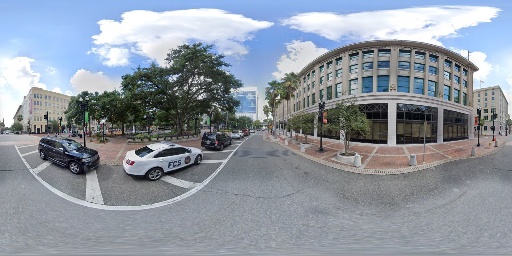}
  \end{subfigure}
    \begin{subfigure}[t]{\sota_width\linewidth}
  \includegraphics[width=\linewidth]{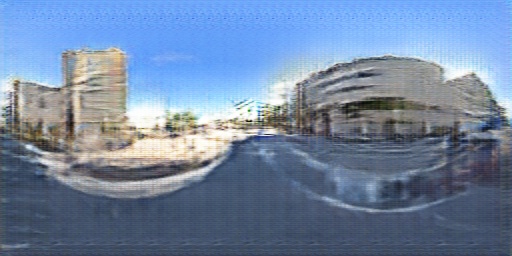}
  \end{subfigure}
  \begin{subfigure}[t]{\sota_width\linewidth}
  \includegraphics[width=\linewidth]{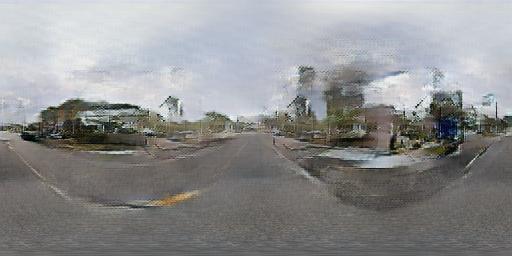}
  \end{subfigure}
  \begin{subfigure}[t]{\sota_width\linewidth}
    \includegraphics[width=\linewidth]{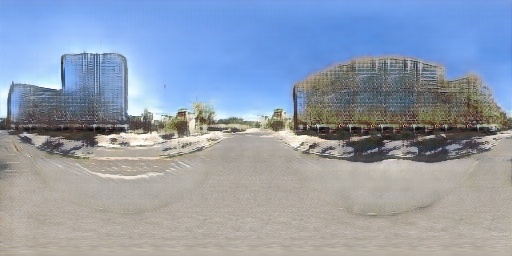}
  \end{subfigure}
  \begin{subfigure}[t]{\sota_width\linewidth}
    \includegraphics[width=\linewidth]{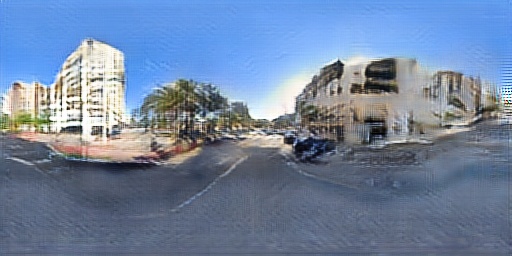}
  \end{subfigure}
  \begin{subfigure}[t]{\sota_width\linewidth}
    \includegraphics[width=\linewidth]{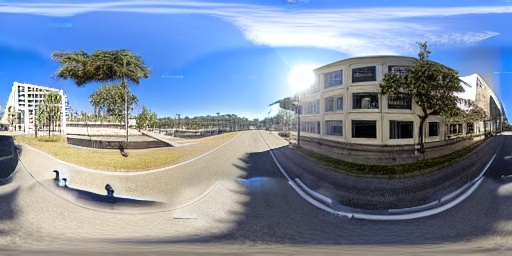}
  \end{subfigure}
  \begin{subfigure}[t]{\sota_width\linewidth}
  \includegraphics[width=\linewidth]{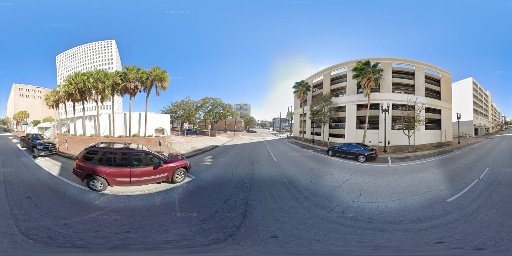}
  \end{subfigure}
    \begin{subfigure}[t]{\sota_width\linewidth}
  \includegraphics[width=\linewidth]{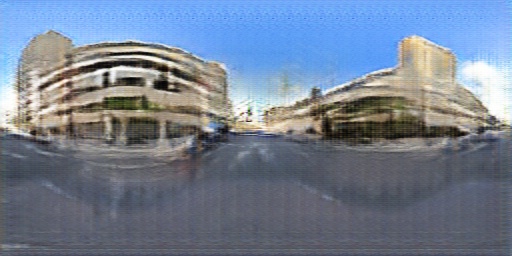}
  \end{subfigure}
  \begin{subfigure}[t]{\sota_width\linewidth}
  \includegraphics[width=\linewidth]{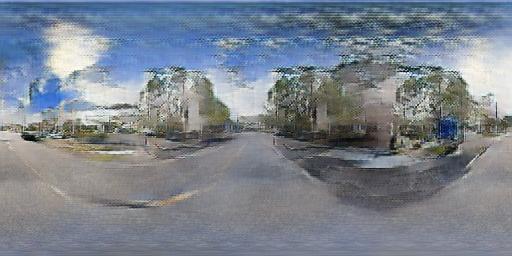}
  \end{subfigure}
  \begin{subfigure}[t]{\sota_width\linewidth}
    \includegraphics[width=\linewidth]{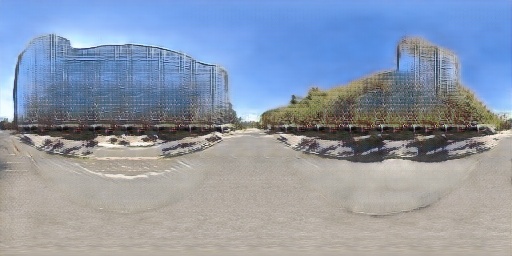}
  \end{subfigure}
  \begin{subfigure}[t]{\sota_width\linewidth}
    \includegraphics[width=\linewidth]{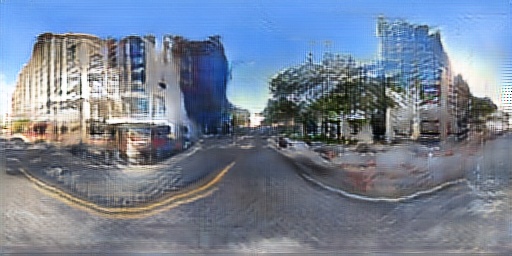}
  \end{subfigure}
  \begin{subfigure}[t]{\sota_width\linewidth}
    \includegraphics[width=\linewidth]{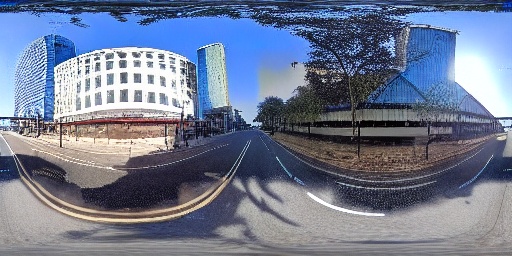}
  \end{subfigure}
  \begin{subfigure}[t]{\sota_width\linewidth}
  \includegraphics[width=\linewidth]{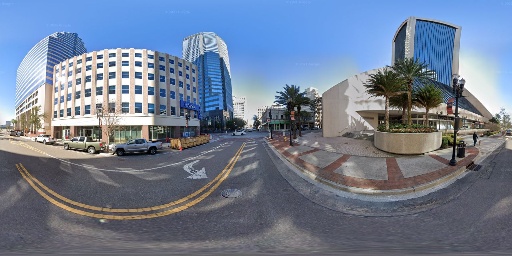}
  \end{subfigure}
    \begin{subfigure}[t]{\sota_width\linewidth}
  \includegraphics[width=\linewidth]{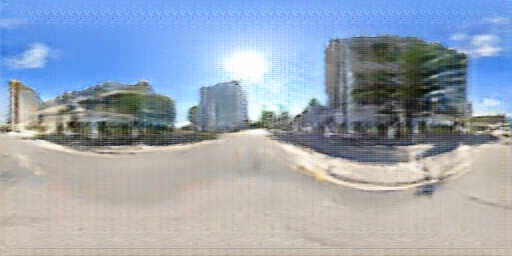}
  \end{subfigure}
  \begin{subfigure}[t]{\sota_width\linewidth}
  \includegraphics[width=\linewidth]{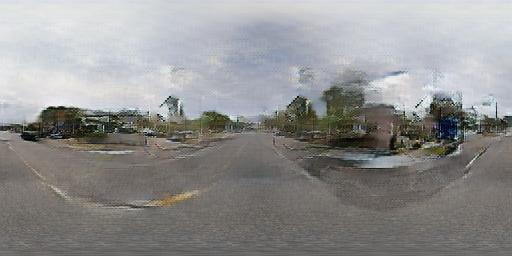}
  \end{subfigure}
  \begin{subfigure}[t]{\sota_width\linewidth}
    \includegraphics[width=\linewidth]{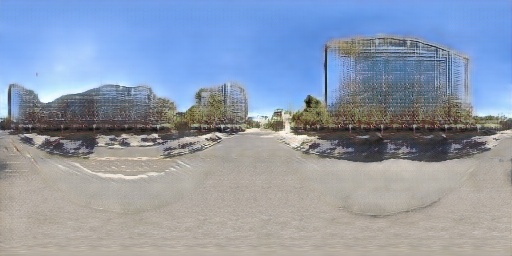}
  \end{subfigure}
  \begin{subfigure}[t]{\sota_width\linewidth}
    \includegraphics[width=\linewidth]{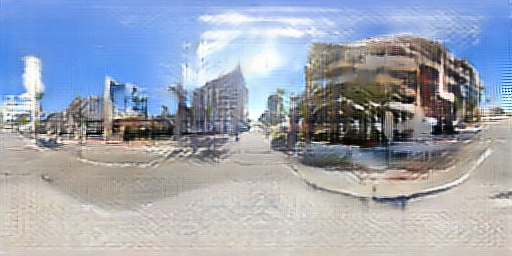}
  \end{subfigure}
  \begin{subfigure}[t]{\sota_width\linewidth}
    \includegraphics[width=\linewidth]{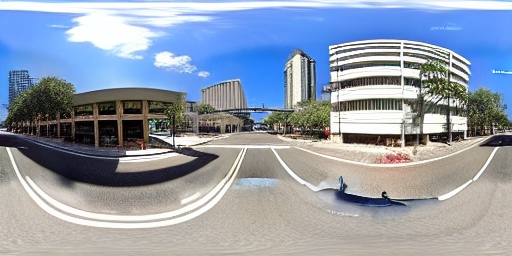}
  \end{subfigure}
  \begin{subfigure}[t]{\sota_width\linewidth}
  \includegraphics[width=\linewidth]{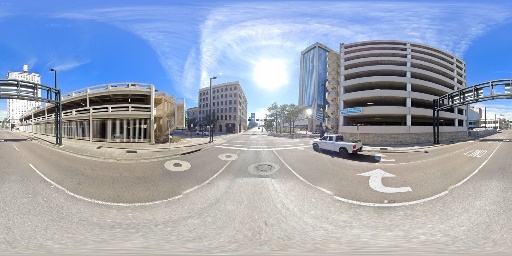}
  \end{subfigure}
    \begin{subfigure}[t]{\sota_width\linewidth}
  \includegraphics[width=\linewidth]{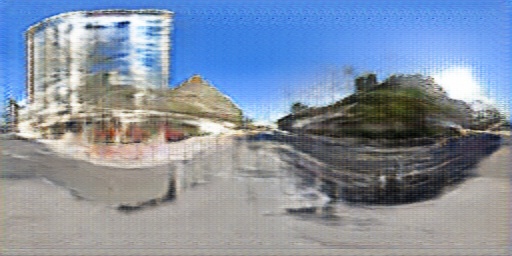}
  \end{subfigure}
  \begin{subfigure}[t]{\sota_width\linewidth}
  \includegraphics[width=\linewidth]{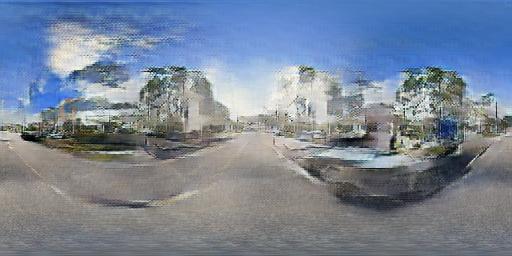}
  \end{subfigure}
  \begin{subfigure}[t]{\sota_width\linewidth}
    \includegraphics[width=\linewidth]{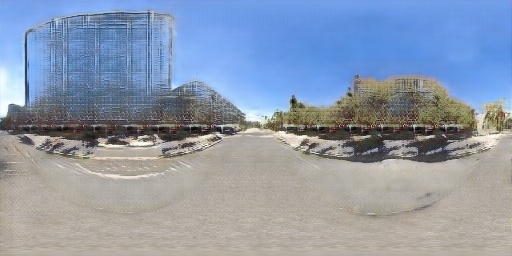}
  \end{subfigure}
  \begin{subfigure}[t]{\sota_width\linewidth}
    \includegraphics[width=\linewidth]{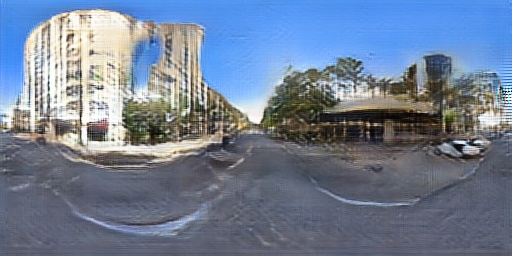}
  \end{subfigure}
  \begin{subfigure}[t]{\sota_width\linewidth}
    \includegraphics[width=\linewidth]{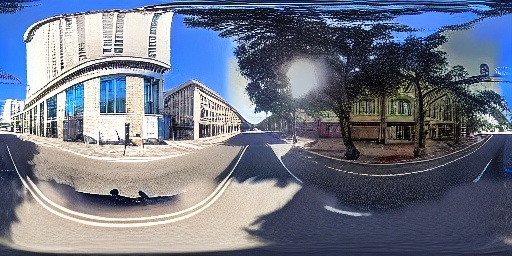}
  \end{subfigure}
  \begin{subfigure}[t]{\sota_width\linewidth}
  \includegraphics[width=\linewidth]{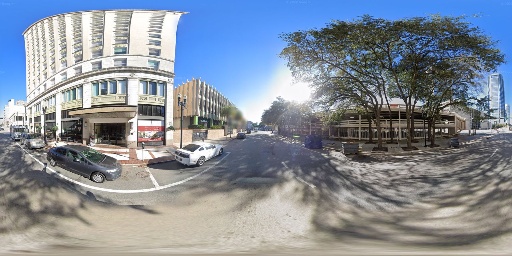}
  \end{subfigure}
    \begin{subfigure}[t]{\sota_width\linewidth}
  \includegraphics[width=\linewidth]{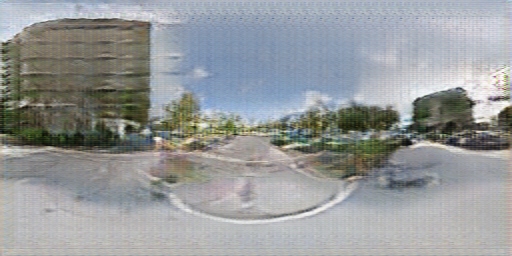}
  \end{subfigure}
  \begin{subfigure}[t]{\sota_width\linewidth}
  \includegraphics[width=\linewidth]{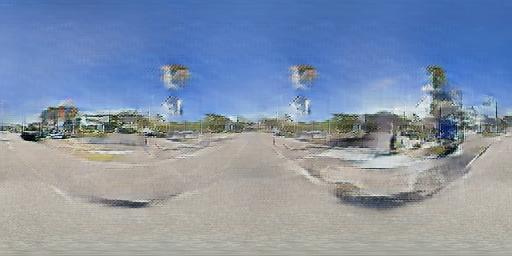}
  \end{subfigure}
  \begin{subfigure}[t]{\sota_width\linewidth}
    \includegraphics[width=\linewidth]{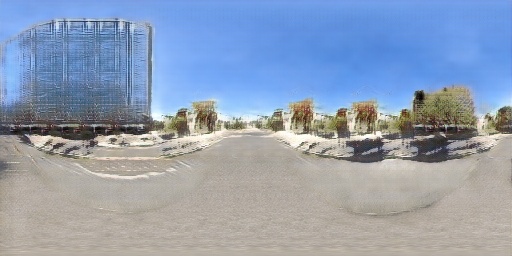}
  \end{subfigure}
  \begin{subfigure}[t]{\sota_width\linewidth}
    \includegraphics[width=\linewidth]{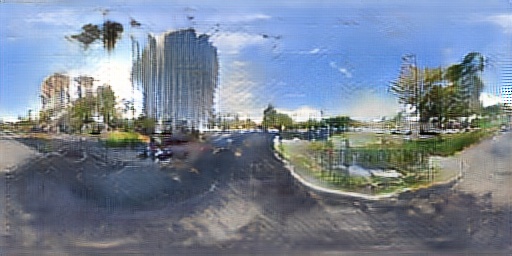}
  \end{subfigure}
  \begin{subfigure}[t]{\sota_width\linewidth}
    \includegraphics[width=\linewidth]{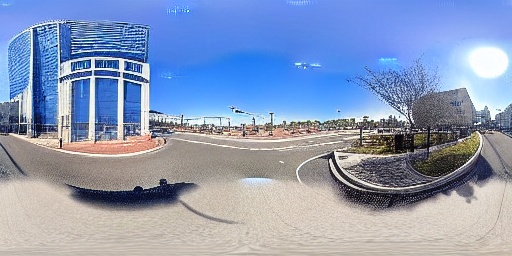}
  \end{subfigure}
  \begin{subfigure}[t]{\sota_width\linewidth}
  \includegraphics[width=\linewidth]{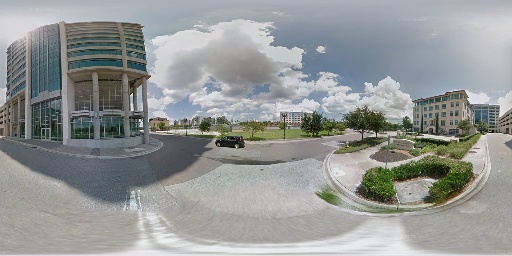}
  \end{subfigure}
    \begin{subfigure}[t]{\sota_width\linewidth}
  \includegraphics[width=\linewidth]{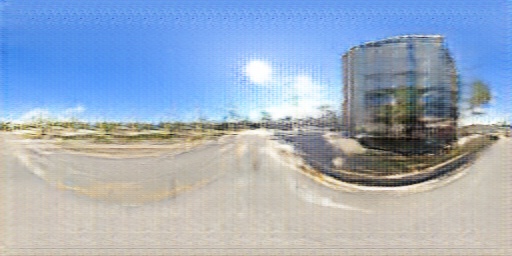}
    \caption{Sat2G}
  \end{subfigure}
  \begin{subfigure}[t]{\sota_width\linewidth}
  \includegraphics[width=\linewidth]{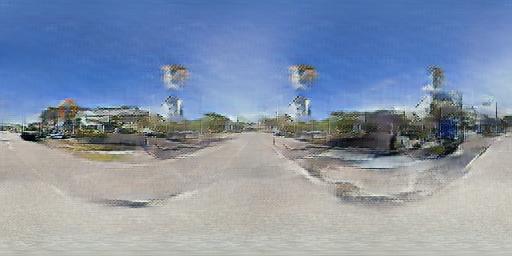}
    \caption{Sat2D}
  \end{subfigure}
  \begin{subfigure}[t]{\sota_width\linewidth}
    \includegraphics[width=\linewidth]{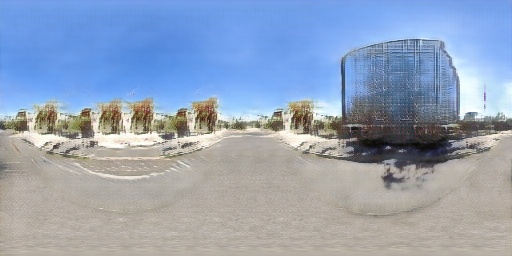}
      \caption{CrossMLP}
  \end{subfigure}
  \begin{subfigure}[t]{\sota_width\linewidth}
    \includegraphics[width=\linewidth]{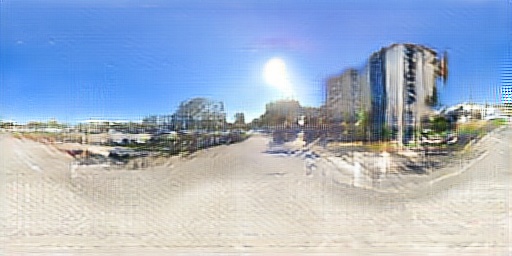}
      \caption{PanoGAN}
  \end{subfigure}
  \begin{subfigure}[t]{\sota_width\linewidth}
    \includegraphics[width=\linewidth]{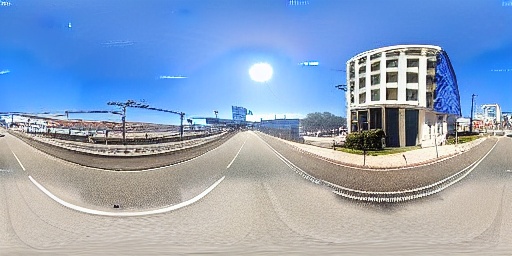}
        \caption{Ours}
  \end{subfigure}
  \begin{subfigure}[t]{\sota_width\linewidth}
  \includegraphics[width=\linewidth]{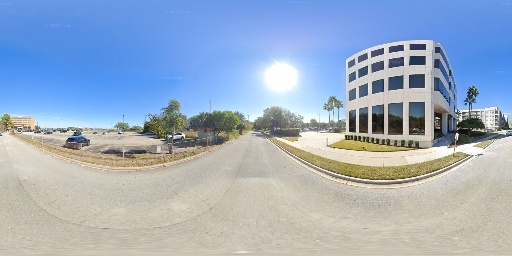}
\caption{GT}
  \end{subfigure}
\caption{Qualitative comparison. We present various synthesis results of our method, compared with Sat2Ground \cite{lu2020geometry}, Sat2Density \cite{qian2023sat2density}, CrossMLP \cite{ren2021cascaded}, PanoGAN \cite{wu2022cross}. Our results are more photorealistic than the baseline methods.}
\label{fig:sota_qualit}
\end{figure}

\textbf{Baseline methods}. We chose two direct synthesis methods, \textbf{CrossMLP} \cite{ren2021cascaded} and \textbf{PanoGAN} \cite{wu2022cross} and two geometry-guided methods \textbf{Sat2Ground} \cite{lu2020geometry} and \textbf{Sat2Density} \cite{qian2023sat2density} as baseline methods. The implementation and modification can be found in supplementary material. 

\textbf{Evaluation metrics}. We use a combination of low-level, structure-level, semantic-level, and perceptual-level metrics to evaluate the quality of synthesized results. \textbf{Low-level metrics}. We follow \cite{wu2022cross,regmi2018cross,lu2020geometry} and use PSNR, and SSIM, which evaluate differences per pixels or local patches. \textbf{Edge-level metrics}. We extract the edge map from the synthesized result and ground truth using the Canny detector and calculate their average IoU (intersection over union) as the edge level similarity metric, denoted as \(I_e\). \textbf{Semantic level metrics}: We calculate the average IoU of building (\(I_B\)), ground (\(I_G\)), and sky labels (\(I_S\)) between synthesized and ground truth images, where the semantic labels are generated from OneFormer \cite{jain2023oneformer} trained on ADE20K dataset \cite{zhou2017scene}. \textbf{Perceptual level metrics}: we apply three widely-used perceptual metrics: the Fréchet Inception Distance (FID) \cite{heusel2017gans} and the Learned Perceptual Image Patch Similarity (LPIPS) \cite{zhang2018unreasonable}, and DreamSIM \cite{fu2023dreamsim}.

\subsection{Comparison to State-of-the-Art Methods}

\cref{tab:sota_quant} and \cref{fig:sota_qualit} provide quantitative and qualitative comparison results of CrossMLP \cite{ren2021cascaded}, PanoGAN \cite{wu2022cross}, Sat2Ground \cite{lu2020geometry}, Sat2Density \cite{qian2023sat2density} and ours in the dataset described in \cref{sec:dataset}.

\textbf{Semantic level similarity}. For a semantic category, a higher mIoU indicates the synthesized objects are more recognized by the pre-trained semantic segmentation models. For building objects, ours achieved 25.8\% and 54.8\% improvement than CrossMLP and PanoGAN. Similarly for road objects, ours achieved 685\% and 402\% improvement than CrossMLP and PanoGAN. Specifically, Sat2Ground synthesized the basic building layouts because of its geometry-guided module while the synthesized building objects were hardly recognized by the pre-trained model, which can be attributed to its GAN-based synthsis module. Moreover, the road regions generated by CrossMLP and PanoGAN exhibit blurry and repetitive artifacts, which are misclassified as 'Sand' and 'Earth' categories, and a similar issue occurs with the "Building" label.

\textbf{Perceptual level similarity}. As shown in \cref{fig:sota_qualit}, our approach produced fewer artifacts, and the synthesized building facades were more similar to those in the ground truth images. As shown in \cref{tab:sota_quant}, ours achieved better FID results with more than 15.7\% compared to baseline methods. The baseline methods generated artifacts of varying degrees around the building and ground areas. CrossMLP, in particular, synthesized buildings with transparent effects that blend with the sky and vegetation. PanoGAN and Sat2Ground performed slightly better than CrossMLP around building regions, displaying the basic facade layouts. However, these layouts appear to be more randomly generated and contain numerous repetitive artifacts.

\textbf{Edge and low-level similarity}. For edge-level performance, ours achieved an improvement of more than 15.5\% than baseline methods. This superiority was attributed to the ground-view satellite texture conditions offering more high-frequency information compared to semantics. For low-level performance, ours ranked second in SSIM and third in PSNR. Given that these metrics evaluate differences at the per-pixel or patch level, our diffusion-based synthesis model is not designed to precisely replicate dataset distributions at the pixel level. Instead, it focuses on synthesizing structures and perceptual features. Additionally, these metrics are particularly sensitive to the inherent randomness of diffusion-based models, which arises from their training on large-scale datasets. For instance, in \cref{fig:sota_qualit}, rows 3, 6, 7, and 9 of our synthesized results exhibit significant deviations from the ground truth at the pixel level, including variations in lighting, shadows, clouds, and tree shapes.

\subsection{Ablation Study}
\begin{table}[tb]
  \caption{Quantitative ablation study of the proposed geospecific priors, the ground-view satellite texture condition, termed as "RGB". The same settings for \cref{fig:ablation_ours}.}
  \centering
\begin{tabular}{c c c c c c c c c c c c}
    \toprule
    \textbf{Method} & \multicolumn{2}{c}{\textbf{Low level}} & \textbf{Edge level} &\multicolumn{3}{c}{\textbf{Semantic level}} & \multicolumn{3}{c}{\textbf{Perceptual Level}} \\
 & PSNR↑ & SSIM↑ & \(I_E\)↑ & \(I_B\)↑ & \(I_G\)↑ & \(I_S\)↑ & LPIPS↓ & FID↓ & DreamSIM↓\\ 
    \midrule
    Ours & \textbf{19.82} & \textbf{0.389} & \textbf{0.090} & \textbf{0.550} & \textbf{0.898} & \textbf{0.876} & \textbf{0.456} & \textbf{66.06} & \textbf{0.309}\\ 
 w/o prior & 17.34 & 0.288 & 0.071 & 0.359 & 0.832 & 0.726 & 0.580 & 87.75 & 0.416 \\ 
  w/o RGB & 18.74 & 0.350 & 0.086  & 0.521 & 0.722 & 0.862 & 0.587 & 89.60 & 0.324\\  
    \bottomrule
  \end{tabular}
  \label{tab:quan_ablation}
\end{table}

\begin{figure}[tb]
    \centering
  \begin{subfigure}[t]{0.24\linewidth}
      \includegraphics[width=\linewidth]{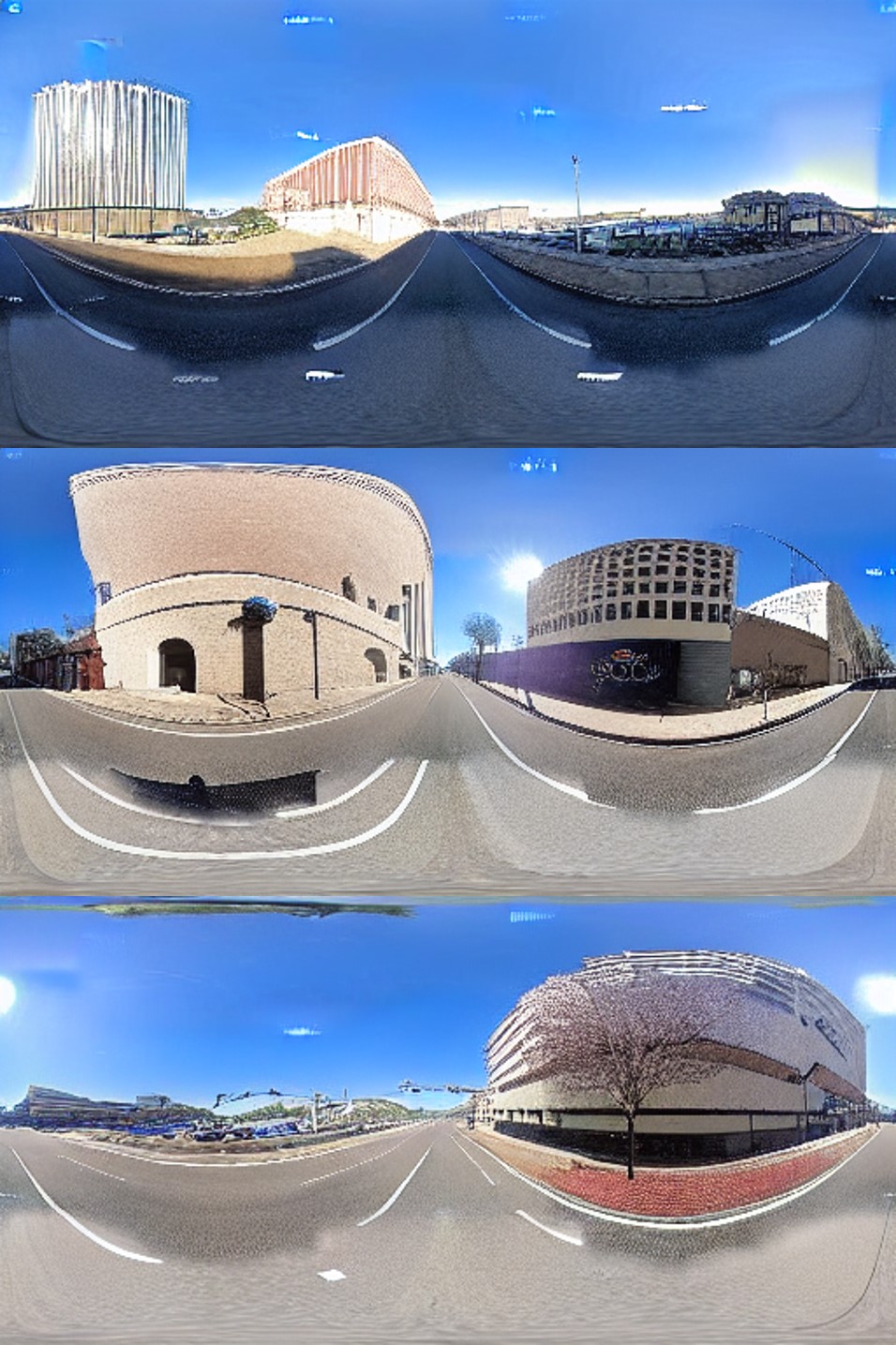}
        \caption{Ours w/o RGB}
  \end{subfigure}
    \begin{subfigure}[t]{0.24\linewidth}
      \includegraphics[width=\linewidth]{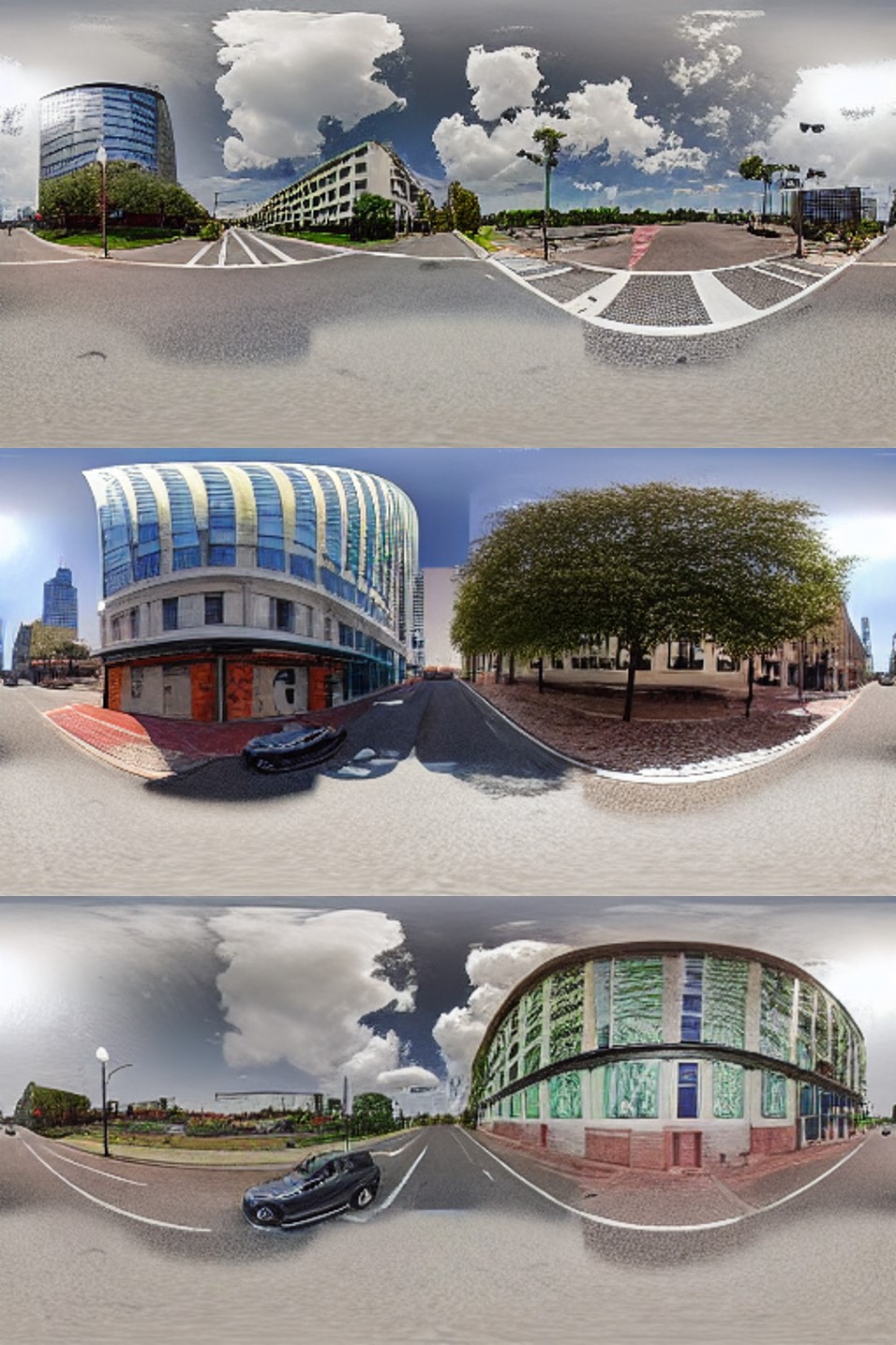}
        \caption{Ours w/o prior}
  \end{subfigure}
    \begin{subfigure}[t]{0.24\linewidth}
      \includegraphics[width=\linewidth]{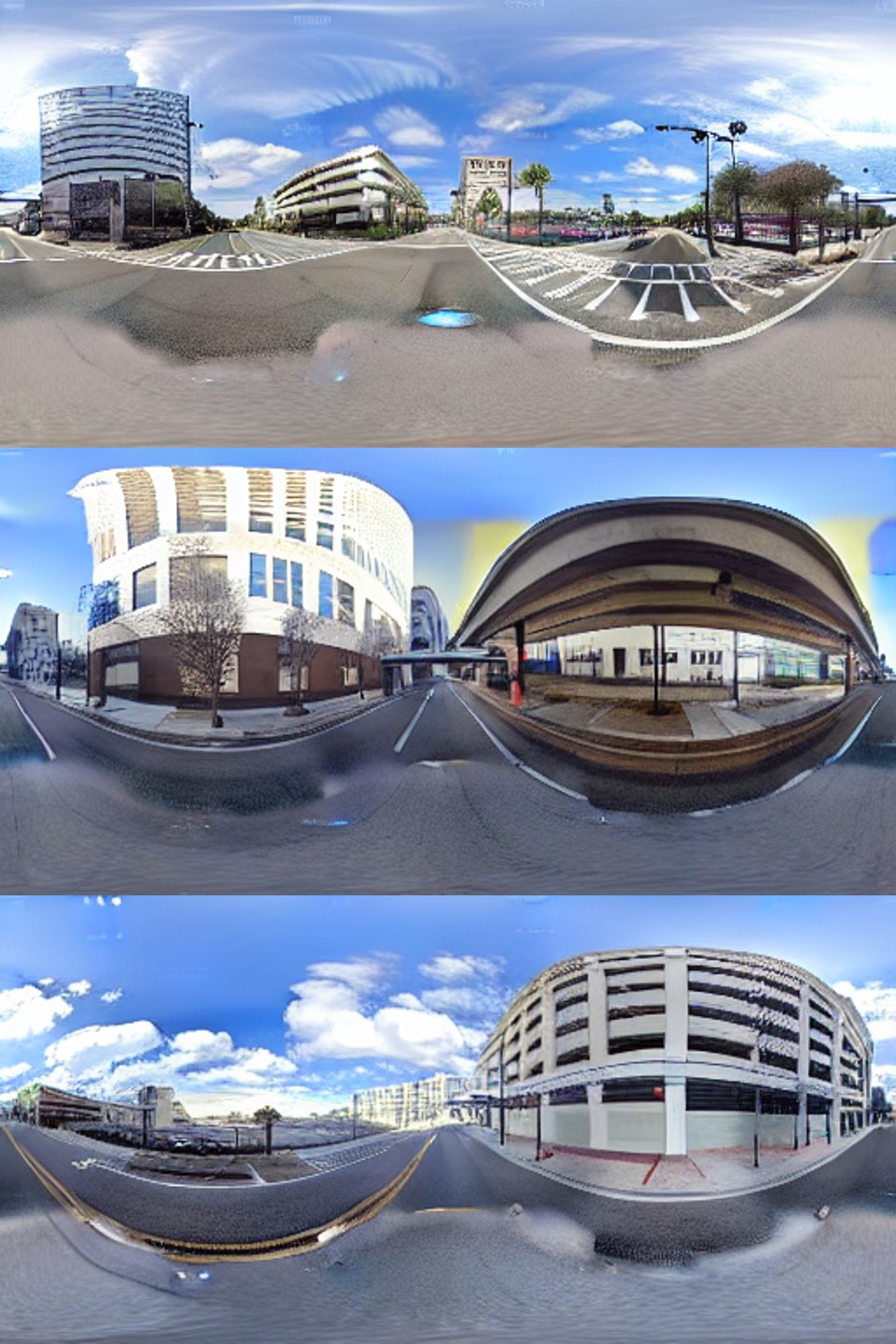}
        \caption{Ours}
  \end{subfigure}
    \begin{subfigure}[t]{0.24\linewidth}
      \includegraphics[width=\linewidth]{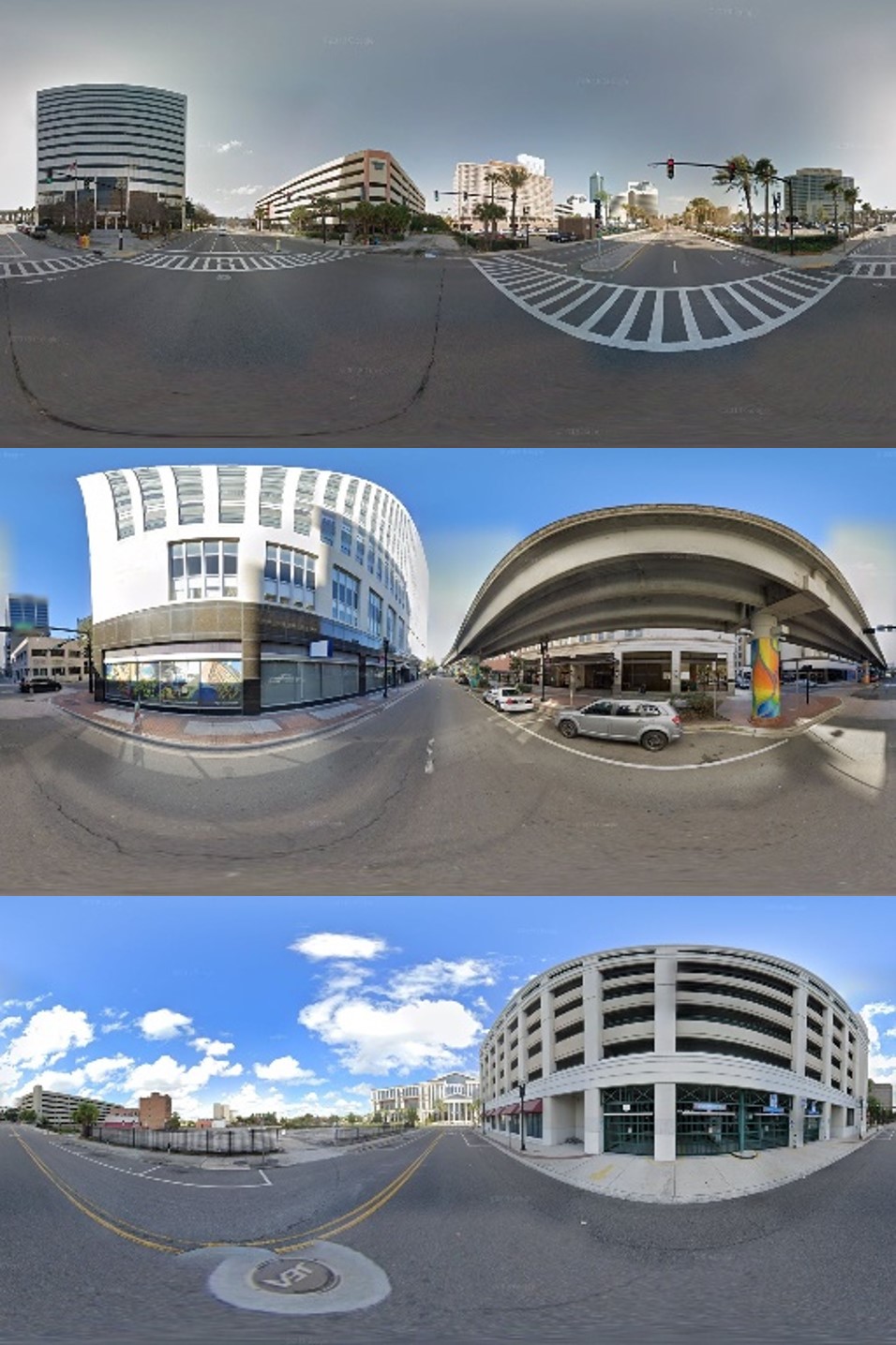}
        \caption{GT}
  \end{subfigure}
    \caption{Qualitative ablation study. We show the visual comparison of the synthesized images by our methods with different configurations.}
    \label{fig:ablation_ours}
\end{figure}



We further investigate the influence of multiple key components of our pipelines.


\textbf{Importance of geo-lcoation prior}. We removed the geospecific prior, formulated as the additional parameters in the cross-attention module of the diffusion model, and performed the training to see the influence of geospecific prior. As shown in \cref{tab:quan_ablation}, ours using geospecific prior achieved the improvement of 32.8\% and 25.7\% in FID and DreamSIM, 53.2\% in \(mIoU\) of the "Building" label than ours without prior.

\textbf{Importance of texture condition}.We substituted ground-view satellite textures with semantics and trained our models to assess the significance of texture conditions. As shown in \cref{tab:quan_ablation}, our approach utilizing texture conditions surpassed the one employing semantics by 23.5\%, 20.7\%, and 4.6\% across three perceptual-level metrics. The ground-view satellite texture provides richer details, thereby establishing a less ambiguous mapping relation. Moreover, the qualitative results depicted in \cref{fig:ablation_ours} underscore the challenges faced by our semantics-dependent approach in accurately synthesizing building facade layouts and textures.

\begin{table}[tb]
  \centering
    \caption{Quantitative results for the potential improvement of baseline methods with our ground-view satellite texture conditions over semantics.}
\begin{tabular}{c c c c c c c}
    \toprule
    \textbf{Method} & \multicolumn{2}{c}{\textbf{LPIPS}\(\downarrow\)} &\multicolumn{2}{c}{\textbf{FID}\(\downarrow\)} & \multicolumn{2}{c}{\textbf{DreamSIM}\(\downarrow\)} \\
 & Semantics & Textures & Semantics & Textures & Semantics & Textures\\ 
    \midrule
    CrossMLP\cite{ren2021cascaded} & \textbf{0.509} & 0.535 & 170.8 & \textbf{86.65} & 0.434 & \textbf{0.432} \\
    PanoGAN\cite{wu2022cross} & 0.488 & \textbf{0.477} & 98.81 & \textbf{84.38} & 0.348 & \textbf{0.337} \\
    Ours & 0.587 & \textbf{0.449} & 89.60 & \textbf{71.04} & 0.324 & \textbf{0.315} \\
    \bottomrule
  \end{tabular}
  \label{tab:aba_potential}
\end{table}

\textbf{Potential improvement for baselines}. To showcase the effectiveness of the proposed ground-view satellite textures, we conducted an experiment comparing them with two baseline methods, CrossMLP \cite{ren2021cascaded} and PanoGAN \cite{wu2022cross}, which originally utilize ground-view semantics as auxiliary inputs. Instead of semantics, we integrated ground-view textures and assessed their perceptual performance, as illustrated in \cref{fig:ablation_ours}. Overall, with our ground-view texture integration, both baseline methods demonstrated improved performance. Specifically, PanoGAN exhibited enhancements of 2.2\%, 14.6\%, and 3.1\% in LPIPS, FID, and DreamSIM metrics, respectively. CrossMLP, when augmented with texture conditions, showed superior performance in FID and DreamSIM but slightly inferior results in the LPIPS metric.


\begin{figure}[tb]
    \centering
    \includegraphics[width=0.85\linewidth]{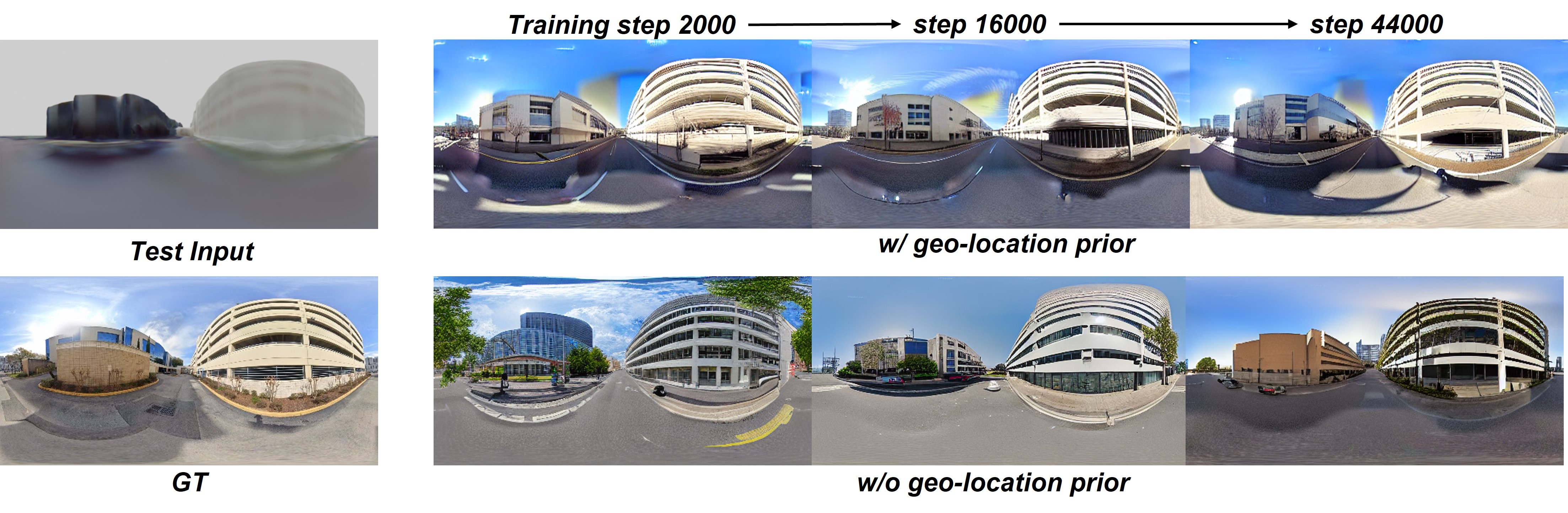}
    \caption{Training efficiency comparison between our method with and without geospecific prior. At a very early stage of the 2000 training step, the geospecific prior enables our method to synthesize similar results to the ground truth.}
    \label{fig:alation_prior}
\end{figure}

\begin{figure}[t]
  \centering
  \includegraphics[width=0.85\linewidth]{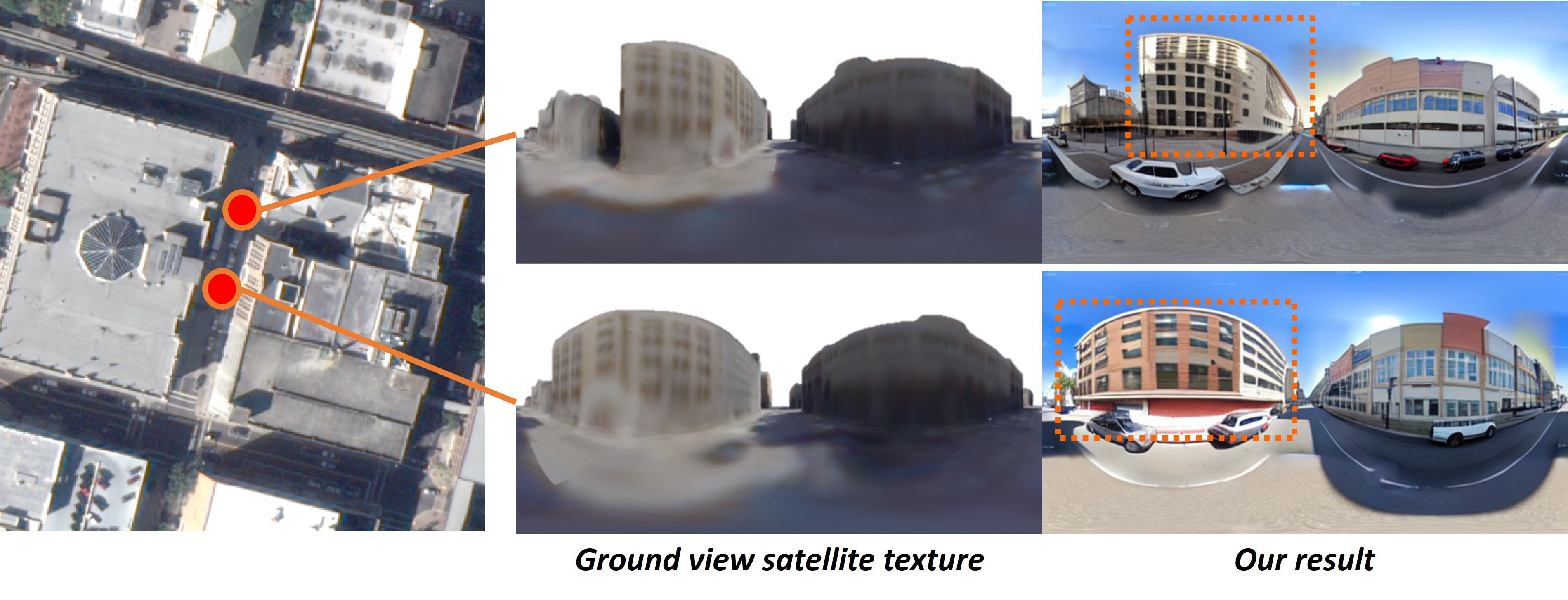}
   \caption{Examples of our limitation. The inherent randomness of the diffusion model makes the synthesis results (marked as orange rectangles) not consistent with their neighbor views.}
   \label{fig:limitation}
\end{figure}

\subsection{Limitation}
Although the synthesized views are geospecific with the help of geospecific prior and ground-view satellite textures, they currently lack view consistency among neighboring views, as shown in \cref{fig:limitation}. For instance, buildings synthesized at two adjacent locations (highlighted by orange rectangles) show inconsistencies in color and facade layouts. This issue arises because the inherent randomness of the diffusion model prevents the results from strictly adhering to the input conditions in the absence of explicit cross-location or view-consistency constraints. Addressing this limitation to produce not only photorealistic but also consistent view sequences is a focus for future work.

\section{Conclusion}
In this paper, we propose a novel pipeline for predicting ground-view images from multi-view satellite images. The predicted ground views are geospecific, in that the generated views are not only consistent with the geometry derived from the satellite views but also the textural information from the satellite view, with a resolution enhancement at a factor of 10 or more. This stands for our work as the first that achieves view prediction that is specifically real to geolocation. In particular, we propose a geometry refinement module to refine satellite 3D geometry, to minimize the texture distortions on building facades, which significantly improves the transfered structural information from the weak satellite textures to the predicted views. Moreover, we propose to use a geospecific prior, to control the learned distributions of the diffusion model, to predict views that respect the local street-view styles. This encoded geospecific prior not only distills the generation to be geospecific but is also shown to be extremely effective in accelerating the training convergence. Our experiments demonstrate that our method significantly outperforms published baselines at a large margin, and is able to predict high-resolution, authentic ground views merely using multi-view satellite images.
\label{sec:conclusion}

\section*{Acknowledgements}
This work is partially supported by the Intelligence Advanced Research Projects Activity (IARPA) via Department of Interior/ Interior Business Center (DOI/IBC) contract number 140D0423C0075. The U.S. Government is authorized to reproduce and distribute reprints for Governmental purposes notwithstanding any copyright annotation thereon. Disclaimer: The views and conclusions contained herein are those of the authors and should not be interpreted as necessarily representing the official policies or endorsements, either expressed or implied, of IARPA, DOI/IBC, or the U.S. Government. It is also supported by the Office of Naval Research (Award No. N000142012141 and N000142312670). The authors would like to thank Xi Liu for his valuable discussion during this work.

%
%
\bibliographystyle{splncs04}
\bibliography{main}

\begin{thebibliography}{10}
\providecommand{\url}[1]{\texttt{#1}}
\providecommand{\urlprefix}{URL }
\providecommand{\doi}[1]{https://doi.org/#1}

\bibitem{cai2019ground}
Cai, S., Guo, Y., Khan, S., Hu, J., Wen, G.: Ground-to-aerial image geo-localization with a hard exemplar reweighting triplet loss. In: Proceedings of the IEEE/CVF International Conference on Computer Vision. pp. 8391--8400 (2019)

\bibitem{castaldo2015semantic}
Castaldo, F., Zamir, A., Angst, R., Palmieri, F., Savarese, S.: Semantic cross-view matching. In: Proceedings of the IEEE International Conference on Computer Vision Workshops. pp. 9--17 (2015)

\bibitem{chan2022learning}
Chan, C., Durand, F., Isola, P.: Learning to generate line drawings that convey geometry and semantics. In: Proceedings of the IEEE/CVF Conference on Computer Vision and Pattern Recognition. pp. 7915--7925 (2022)

\bibitem{choi2018stargan}
Choi, Y., Choi, M., Kim, M., Ha, J.W., Kim, S., Choo, J.: Stargan: Unified generative adversarial networks for multi-domain image-to-image translation. In: Proceedings of the IEEE conference on computer vision and pattern recognition. pp. 8789--8797 (2018)

\bibitem{de2014automatic}
De~Franchis, C., Meinhardt-Llopis, E., Michel, J., Morel, J.M., Facciolo, G.: An automatic and modular stereo pipeline for pushbroom images. ISPRS Annals of the Photogrammetry, Remote Sensing and Spatial Information Sciences  \textbf{2},  49--56 (2014)

\bibitem{Esser_2021_CVPR}
Esser, P., Rombach, R., Ommer, B.: Taming transformers for high-resolution image synthesis. In: Proceedings of the IEEE/CVF Conference on Computer Vision and Pattern Recognition (CVPR). pp. 12873--12883 (June 2021)

\bibitem{fu2023dreamsim}
Fu, S., Tamir, N., Sundaram, S., Chai, L., Zhang, R., Dekel, T., Isola, P.: Dreamsim: Learning new dimensions of human visual similarity using synthetic data. arXiv preprint arXiv:2306.09344  (2023)

\bibitem{heusel2017gans}
Heusel, M., Ramsauer, H., Unterthiner, T., Nessler, B., Hochreiter, S.: Gans trained by a two time-scale update rule converge to a local nash equilibrium. Advances in neural information processing systems  \textbf{30} (2017)

\bibitem{hirschmuller2007stereo}
Hirschmuller, H.: Stereo processing by semiglobal matching and mutual information. IEEE Transactions on pattern analysis and machine intelligence  \textbf{30}(2),  328--341 (2007)

\bibitem{ho2020denoising}
Ho, J., Jain, A., Abbeel, P.: Denoising diffusion probabilistic models. Advances in neural information processing systems  \textbf{33},  6840--6851 (2020)

\bibitem{hu2021lora}
Hu, E.J., Shen, Y., Wallis, P., Allen-Zhu, Z., Li, Y., Wang, S., Wang, L., Chen, W.: Lora: Low-rank adaptation of large language models. arXiv preprint arXiv:2106.09685  (2021)

\bibitem{Hu_2018_CVPR}
Hu, S., Feng, M., Nguyen, R.M.H., Lee, G.H.: Cvm-net: Cross-view matching network for image-based ground-to-aerial geo-localization. In: Proceedings of the IEEE Conference on Computer Vision and Pattern Recognition (CVPR) (June 2018)

\bibitem{huang2022evaluation}
Huang, D., Tang, Y., Qin, R.: An evaluation of planetscope images for 3d reconstruction and change detection--experimental validations with case studies. GIScience \& Remote Sensing  \textbf{59}(1),  744--761 (2022)

\bibitem{isola2017image}
Isola, P., Zhu, J.Y., Zhou, T., Efros, A.A.: Image-to-image translation with conditional adversarial networks. In: Proceedings of the IEEE conference on computer vision and pattern recognition. pp. 1125--1134 (2017)

\bibitem{jain2023oneformer}
Jain, J., Li, J., Chiu, M.T., Hassani, A., Orlov, N., Shi, H.: Oneformer: One transformer to rule universal image segmentation. In: Proceedings of the IEEE/CVF Conference on Computer Vision and Pattern Recognition. pp. 2989--2998 (2023)

\bibitem{lentsch2023slicematch}
Lentsch, T., Xia, Z., Caesar, H., Kooij, J.F.: Slicematch: Geometry-guided aggregation for cross-view pose estimation. In: Proceedings of the IEEE/CVF Conference on Computer Vision and Pattern Recognition. pp. 17225--17234 (2023)

\bibitem{Leotta_2019_CVPR_Workshops}
Leotta, M.J., Long, C., Jacquet, B., Zins, M., Lipsa, D., Shan, J., Xu, B., Li, Z., Zhang, X., Chang, S.F., Purri, M., Xue, J., Dana, K.: Urban semantic 3d reconstruction from multiview satellite imagery. In: Proceedings of the IEEE/CVF Conference on Computer Vision and Pattern Recognition (CVPR) Workshops (June 2019)

\bibitem{li2024sat2scene}
Li, Z., Li, Z., Cui, Z., Pollefeys, M., Oswald, M.R.: Sat2scene: 3d urban scene generation from satellite images with diffusion. arXiv preprint arXiv:2401.10786  (2024)

\bibitem{li2021sat2vid}
Li, Z., Li, Z., Cui, Z., Qin, R., Pollefeys, M., Oswald, M.R.: Sat2vid: street-view panoramic video synthesis from a single satellite image. In: Proceedings of the IEEE/CVF International Conference on Computer Vision. pp. 12436--12445 (2021)

\bibitem{dfc2019}
Lian, Y., Feng, T., Zhou, J., Jia, M., Li, A., Wu, Z., Jiao, L., Brown, M., Hager, G., Yokoya, N., Hänsch, R., Saux, B.L.: Large-scale semantic 3-d reconstruction: Outcome of the 2019 ieee grss data fusion contest—part b. IEEE Journal of Selected Topics in Applied Earth Observations and Remote Sensing  \textbf{14},  1158--1170 (2021). \doi{10.1109/JSTARS.2020.3035274}

\bibitem{lin2013cross}
Lin, T.Y., Belongie, S., Hays, J.: Cross-view image geolocalization. In: Proceedings of the IEEE Conference on Computer Vision and Pattern Recognition. pp. 891--898 (2013)

\bibitem{lu2020geometry}
Lu, X., Li, Z., Cui, Z., Oswald, M.R., Pollefeys, M., Qin, R.: Geometry-aware satellite-to-ground image synthesis for urban areas. In: Proceedings of the IEEE/CVF Conference on Computer Vision and Pattern Recognition. pp. 859--867 (2020)

\bibitem{lugmayr2022repaint}
Lugmayr, A., Danelljan, M., Romero, A., Yu, F., Timofte, R., Van~Gool, L.: Repaint: Inpainting using denoising diffusion probabilistic models. In: Proceedings of the IEEE/CVF Conference on Computer Vision and Pattern Recognition. pp. 11461--11471 (2022)

\bibitem{OpenStreetMap}
{OpenStreetMap contributors}: {Planet dump retrieved from https://planet.osm.org }. \url{ https://www.openstreetmap.org } (2017)

\bibitem{pathak2016context}
Pathak, D., Krahenbuhl, P., Donahue, J., Darrell, T., Efros, A.A.: Context encoders: Feature learning by inpainting. In: Proceedings of the IEEE conference on computer vision and pattern recognition. pp. 2536--2544 (2016)

\bibitem{qian2023sat2density}
Qian, M., Xiong, J., Xia, G.S., Xue, N.: Sat2density: Faithful density learning from satellite-ground image pairs. arXiv preprint arXiv:2303.14672  (2023)

\bibitem{RPC}
Qin, R.: Rpc stereo processor (rsp)--a software package for digital surface model and orthophoto generation from satellite stereo imagery. ISPRS Annals of the Photogrammetry, Remote Sensing and Spatial Information Sciences  \textbf{3},  77--82 (2016)

\bibitem{reed2016learning}
Reed, S.E., Akata, Z., Mohan, S., Tenka, S., Schiele, B., Lee, H.: Learning what and where to draw. Advances in neural information processing systems  \textbf{29} (2016)

\bibitem{regmi2018cross}
Regmi, K., Borji, A.: Cross-view image synthesis using conditional gans. In: Proceedings of the IEEE conference on Computer Vision and Pattern Recognition. pp. 3501--3510 (2018)

\bibitem{regmi2019bridging}
Regmi, K., Shah, M.: Bridging the domain gap for ground-to-aerial image matching. In: Proceedings of the IEEE/CVF International Conference on Computer Vision. pp. 470--479 (2019)

\bibitem{ren2021cascaded}
Ren, B., Tang, H., Sebe, N.: Cascaded cross mlp-mixer gans for cross-view image translation. arXiv preprint arXiv:2110.10183  (2021)

\bibitem{roich2022pivotal}
Roich, D., Mokady, R., Bermano, A.H., Cohen-Or, D.: Pivotal tuning for latent-based editing of real images. ACM Transactions on Graphics (TOG)  \textbf{42}(1),  1--13 (2022)

\bibitem{rombach2022high}
Rombach, R., Blattmann, A., Lorenz, D., Esser, P., Ommer, B.: High-resolution image synthesis with latent diffusion models. In: Proceedings of the IEEE/CVF conference on computer vision and pattern recognition. pp. 10684--10695 (2022)

\bibitem{ruiz2022dreambooth}
Ruiz, N., Li, Y., Jampani, V., Pritch, Y., Rubinstein, M., Aberman, K.: Dreambooth: Fine tuning text-to-image diffusion models for subject-driven generation. arXiv preprint arXiv:2208.12242  (2022)

\bibitem{ruiz2023dreambooth}
Ruiz, N., Li, Y., Jampani, V., Pritch, Y., Rubinstein, M., Aberman, K.: Dreambooth: Fine tuning text-to-image diffusion models for subject-driven generation. In: Proceedings of the IEEE/CVF Conference on Computer Vision and Pattern Recognition. pp. 22500--22510 (2023)

\bibitem{shi2023boosting}
Shi, Y., Wu, F., Perincherry, A., Vora, A., Li, H.: Boosting 3-dof ground-to-satellite camera localization accuracy via geometry-guided cross-view transformer. In: Proceedings of the IEEE/CVF International Conference on Computer Vision. pp. 21516--21526 (2023)

\bibitem{singh2008rational}
Singh, S.K., Naidu, S.D., Srinivasan, T., Krishna, B.G., Srivastava, P.: Rational polynomial modelling for cartosat-1 data. The International Archives of the Photogrammetry, Remote Sensing and Spatial Information Sciences  \textbf{37}(Part B1),  885--888 (2008)

\bibitem{song2020score}
Song, Y., Sohl-Dickstein, J., Kingma, D.P., Kumar, A., Ermon, S., Poole, B.: Score-based generative modeling through stochastic differential equations. arXiv preprint arXiv:2011.13456  (2020)

\bibitem{vo2016localizing}
Vo, N.N., Hays, J.: Localizing and orienting street views using overhead imagery. In: Computer Vision--ECCV 2016: 14th European Conference, Amsterdam, The Netherlands, October 11--14, 2016, Proceedings, Part I 14. pp. 494--509. Springer (2016)

\bibitem{workman2015wide}
Workman, S., Souvenir, R., Jacobs, N.: Wide-area image geolocalization with aerial reference imagery. In: Proceedings of the IEEE International Conference on Computer Vision. pp. 3961--3969 (2015)

\bibitem{wu2022cross}
Wu, S., Tang, H., Jing, X.Y., Zhao, H., Qian, J., Sebe, N., Yan, Y.: Cross-view panorama image synthesis. IEEE Transactions on Multimedia  (2022)

\bibitem{xie2021segformer}
Xie, E., Wang, W., Yu, Z., Anandkumar, A., Alvarez, J.M., Luo, P.: Segformer: Simple and efficient design for semantic segmentation with transformers. Advances in Neural Information Processing Systems  \textbf{34},  12077--12090 (2021)

\bibitem{xu2024large}
Xu, N., Qin, R.: Large-scale dsm registration via motion averaging. ISPRS Annals of the Photogrammetry, Remote Sensing and Spatial Information Sciences  \textbf{10},  275--282 (2024)

\bibitem{xu2024multi}
Xu, N., Qin, R., Huang, D., Remondino, F.: Multi-tiling neural radiance field (nerf)—geometric assessment on large-scale aerial datasets. The Photogrammetric Record  (2024)

\bibitem{xu2023point}
Xu, N., Qin, R., Song, S.: Point cloud registration for lidar and photogrammetric data: A critical synthesis and performance analysis on classic and deep learning algorithms. ISPRS open journal of photogrammetry and remote sensing  \textbf{8},  100032 (2023)

\bibitem{zhang2023adding}
Zhang, L., Rao, A., Agrawala, M.: Adding conditional control to text-to-image diffusion models. In: Proceedings of the IEEE/CVF International Conference on Computer Vision. pp. 3836--3847 (2023)

\bibitem{zhang2018unreasonable}
Zhang, R., Isola, P., Efros, A.A., Shechtman, E., Wang, O.: The unreasonable effectiveness of deep features as a perceptual metric. In: Proceedings of the IEEE conference on computer vision and pattern recognition. pp. 586--595 (2018)

\bibitem{zhou2017scene}
Zhou, B., Zhao, H., Puig, X., Fidler, S., Barriuso, A., Torralba, A.: Scene parsing through ade20k dataset. In: Proceedings of the IEEE conference on computer vision and pattern recognition. pp. 633--641 (2017)

\bibitem{zhu2017unpaired}
Zhu, J.Y., Park, T., Isola, P., Efros, A.A.: Unpaired image-to-image translation using cycle-consistent adversarial networks. In: Proceedings of the IEEE international conference on computer vision. pp. 2223--2232 (2017)

\bibitem{zhu2021vigor}
Zhu, S., Yang, T., Chen, C.: Vigor: Cross-view image geo-localization beyond one-to-one retrieval. In: Proceedings of the IEEE/CVF Conference on Computer Vision and Pattern Recognition. pp. 3640--3649 (2021)

\end{thebibliography}
\end{document}